\documentclass[sigconf]{acmart}
\usepackage{bbding}
\usepackage{pifont}
\usepackage{multirow}
\usepackage{amsmath}
\usepackage{graphicx}
\def\eg{\emph{e.g.}}
\def\ie{\emph{i.e.}}
\AtBeginDocument{%
  \providecommand\BibTeX{{%
    \normalfont B\kern-0.5em{\scshape i\kern-0.25em b}\kern-0.8em\TeX}}}

\setcopyright{acmlicensed}
\copyrightyear{2018}
\acmYear{2018}
\acmDOI{XXXXXXX.XXXXXXX}

\acmConference[MM'24]{Make sure to enter the correct
  conference title from your rights confirmation email}{October 28 - November 1,
  2024}{Melbourne, Australia.}
\acmISBN{978-1-4503-XXXX-X/18/06}

\begin{document}

\title{IPAD: Industrial Process Anomaly Detection Dataset}


\author{Jinfan Liu}
\affiliation{%
  \institution{Shanghai Jiao Tong University}
  % \streetaddress{1 Th{\o}rv{\"a}ld Circle}
  \city{Shanghai}
  \country{China}}
\email{ljflnjz@sjtu.edu.cn}

\author{Yichao Yan}
\affiliation{%
  \institution{Shanghai Jiao Tong University}
  % \streetaddress{1 Th{\o}rv{\"a}ld Circle}
  \city{Shanghai}
  \country{China}}
\email{yanyichao@sjtu.edu.cn}

\author{Junjie Li}
\affiliation{%
  \institution{Shanghai Jiao Tong University}
  % \streetaddress{1 Th{\o}rv{\"a}ld Circle}
  \city{Shanghai}
  \country{China}}
\email{junjieli00@sjtu.edu.cn}

\author{Weiming Zhao}
\affiliation{%
  \institution{Shanghai Jiao Tong University}
  % \streetaddress{1 Th{\o}rv{\"a}ld Circle}
  \city{Shanghai}
  \country{China}}
\email{weiming.zhao@sjtu.edu.cn}

\author{Pengzhi Chu}
\affiliation{%
  \institution{Shanghai Jiao Tong University}
  % \streetaddress{1 Th{\o}rv{\"a}ld Circle}
  \city{Shanghai}
  \country{China}}
\email{pzchu@sjtu.edu.cn}

\author{Xingdong Sheng}
\affiliation{%
  \institution{Lenovo Research}
  % \streetaddress{1 Th{\o}rv{\"a}ld Circle}
  \city{Shanghai}
  \country{China}}
\email{shengxd1@lenovo.com}

\author{Yunhui Liu}
\affiliation{%
  \institution{Lenovo Research}
  % \streetaddress{1 Th{\o}rv{\"a}ld Circle}
  \city{Shanghai}
  \country{China}}
\email{liuyhp@lenovo.com}

\author{Xiaokang Yang}
\affiliation{%
  \institution{Shanghai Jiao Tong University}
  % \streetaddress{1 Th{\o}rv{\"a}ld Circle}
  \city{Shanghai}
  \country{China}}
\email{xkyang@sjtu.edu.cn}
\renewcommand{\shortauthors}{author name and author name, et al.}

\begin{abstract}

  Video anomaly detection (VAD) is a challenging task aiming to recognize anomalies in video frames, and existing large-scale VAD researches primarily focus on road traffic and human activity scenes. In industrial scenes, there are often a variety of unpredictable anomalies, and the VAD method can play a significant role in these scenarios. However, there is a lack of applicable datasets and methods specifically tailored for industrial production scenarios due to concerns regarding privacy and security. To bridge this gap, we propose a new dataset, \textbf{IPAD}, specifically designed for VAD in industrial scenarios. The industrial processes in our dataset are chosen through on-site factory research and discussions with engineers. This dataset covers 16 different industrial devices and contains over 6 hours of both synthetic and real-world video footage. Moreover, we annotate the key feature of the industrial process, \ie, periodicity. Based on the proposed dataset, we introduce a period memory module and a sliding window inspection mechanism to effectively investigate the periodic information in a basic reconstruction model. Our framework leverages LoRA adapter to explore the effective migration of pretrained models, which are initially trained using synthetic data, into real-world scenarios. Our proposed dataset and method will fill the gap in the field of industrial video anomaly detection and drive the process of video understanding tasks as well as smart factory deployment. Project page: \href{https://ljf1113.github.io/IPAD_VAD/}{https://ljf1113.github.io/IPAD\_VAD}.
  \begin{figure}[t]	
    \centering	
	\includegraphics[width=1.0\linewidth]{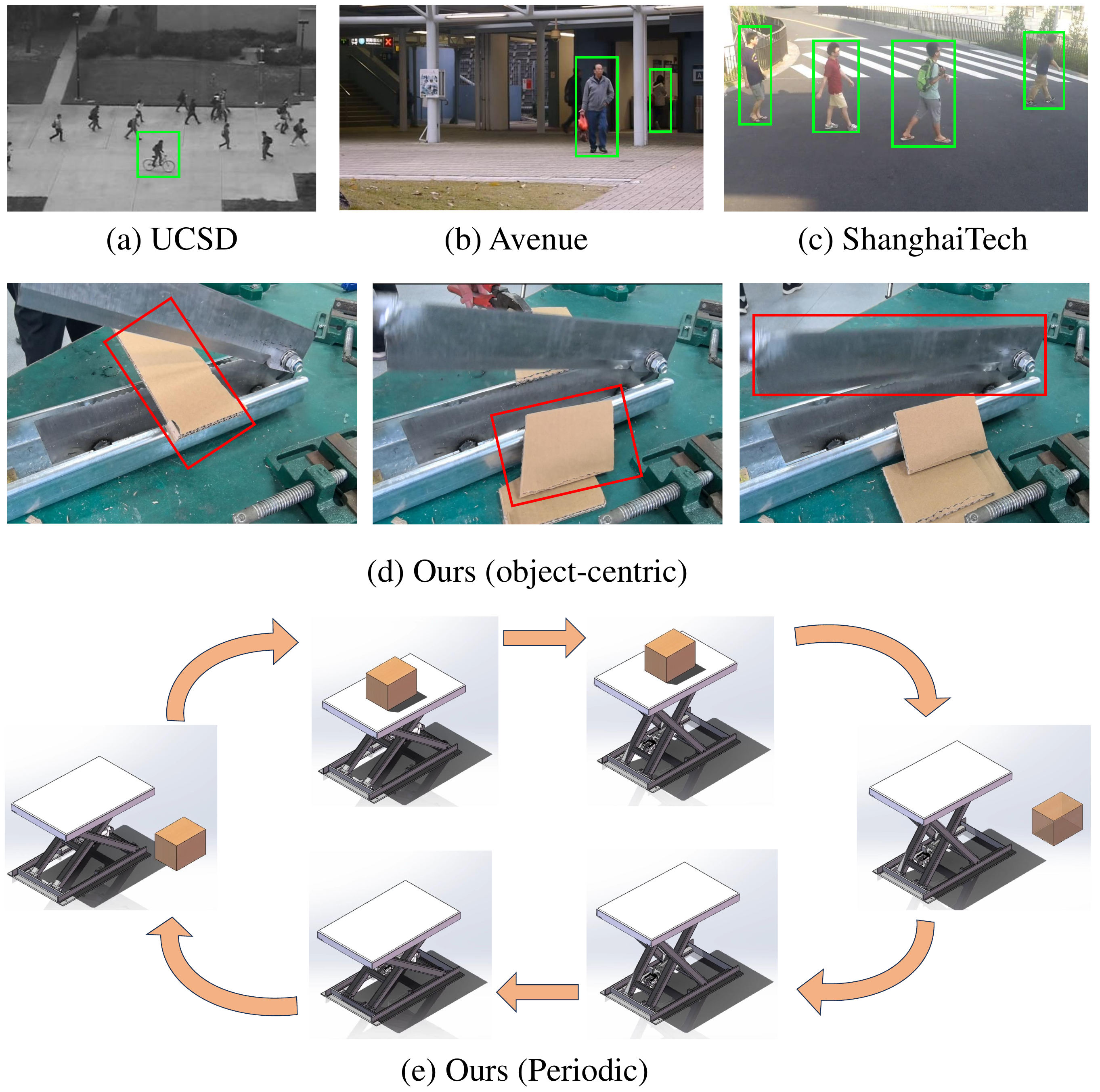}
    \vspace{-6mm}
	\caption{(a)-(c) show the existing human-centric datasets. (d) presents different anomalous regions under the same device in our dataset. (e) shows a periodic action of the automatic lifter in our dataset.
	}
    \label{fig:human-centric}
    % \vspace{-6mm}
\end{figure}
\begin{figure*}[t]	
    \centering	
	\includegraphics[width=1.0\linewidth]{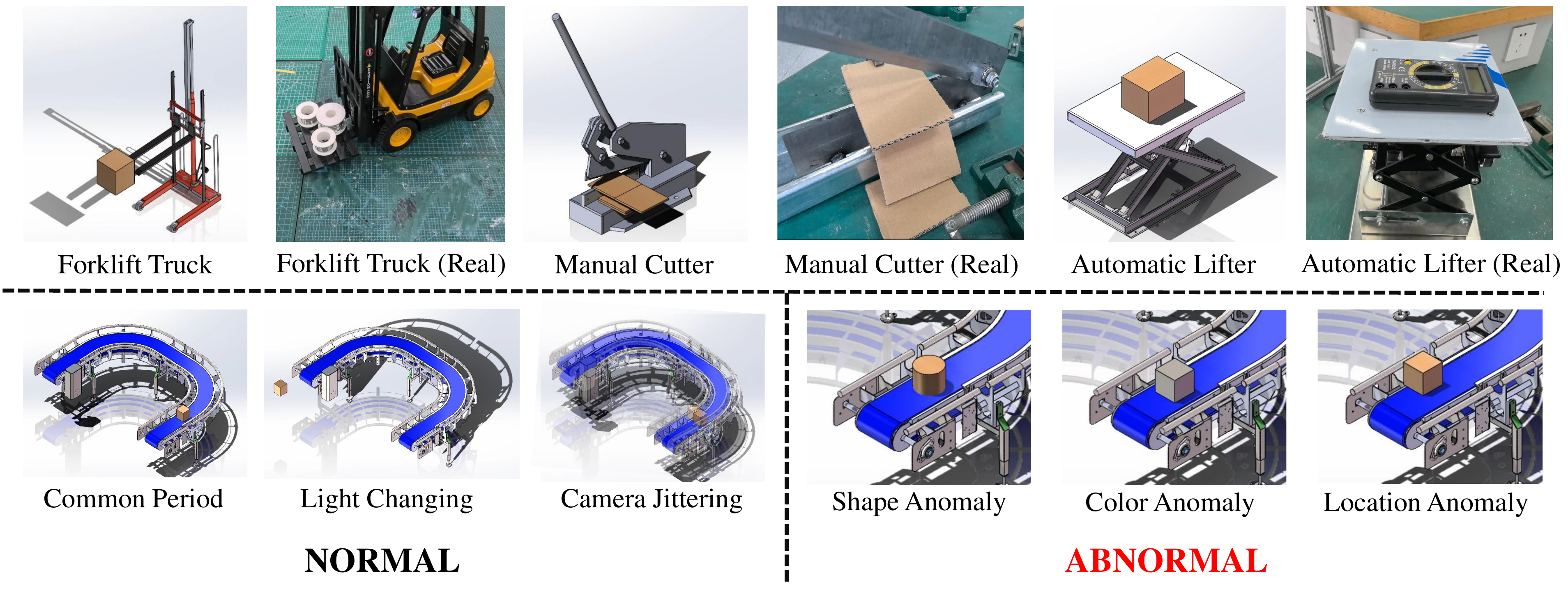}
    \vspace{-6mm}
	\caption{The first line shows part of the devices in our dataset. The second line shows some of the normal and abnormal cases.
	}
    \label{fig:dataset show}
    % \vspace{-2mm}
\end{figure*}
\end{abstract}








\maketitle

\section{Introduction}

\label{sec:intro}

Video Anomaly Detection (VAD) has emerged as a crucial technology in various real-world applications, including traffic safety and intelligent surveillance, due to its remarkable ability to detect multiple anomalous events that may occur in videos. To facilitate the practical implementation of VAD methods, researchers have made significant contributions by introducing rich video datasets and related frameworks tailored to this problem. These datasets cover a wide range of scenarios, with a particular focus on campus and traffic scenes, capturing both normal and abnormal behaviors of pedestrians and vehicles from the surveillance camera perspective~\cite{CUHK, NWPU, ShangHaiTech, Street, Subway, UCSDped, UBnormal, UMN, IITB, zhu2023cross}. The availability of diverse video datasets has paved the way for advanced video anomaly detection approaches. These works have been widely applied in traffic scenarios, including campuses and streets, thereby effectively addressing public safety concerns through the expeditious detection of anomalies~\cite{AE, MemAE, chang2020clustering, li2021variational, liu2018future, liu2021hybrid, lu2020few, sultani2018real, sun2023hierarchical, SVD-GAN, zaheer2020claws, zaigham2021cleaning, zhong2019graph, zou2021progressive, causality-inspired}.

In addition to traffic scenarios, the manufacturing industry is often plagued by a multitude of unpredictable anomalies during equipment operation. Equipment and objects at work can exhibit a wide range of anomalies in appearance and movement, which defy easy categorization, making it challenging to establish a specific framework that encompasses all possible situations. Consequently, substantial human monitoring is required to supervise equipment operations. It would be desirable if we could integrate a video anomaly detection framework into factory surveillance, which could largely alleviate the burden on human monitoring.

However, it is non-trivial to solve this task due to challenges from both data and methodology perspectives. On the one hand, in contrast to the easily accessible road surveillance data, industrial video data is not easy to capture because (1) video information captured within factories would raise concerns regarding \textbf{privacy and security}, making it exceedingly difficult to achieve public access to such data for dataset creation; (2) From an anomaly setting perspective, intentionally making abnormalities can not only damage machinery but also lead to safety accidents.
% Consequently, acquiring ample test data becomes challenging.
% (2) From the perspective of anomaly setting, it is also arduous to artificially generate a substantial number of diverse exception events due to safety considerations.

Another substantial challenge arises in the methodological aspect. Anomaly detection in industrial scenarios differs significantly from tasks in traffic scenarios, both in terms of spatial and temporal considerations. (1) State-of-the-art VAD methods heavily rely on person/object detection results~\cite{Jigsaw,xiao2023divide,reiss2022attribute,hirschorn2023normalizing,barbalau2023ssmtl++}. 
However, accurately detecting each device and object in industrial scenes is unfeasible. Because anomalies in industrial scenes can occur anywhere within the frame, as opposed to being restricted to pedestrian or specific object areas as commonly observed in the traffic scenes, as illustrated in Figure~\ref{fig:human-centric}. (2) At the temporal level, there exists meaningful periodic information regarding the operation process of equipment, which is absent in human-centric scenarios. Existing methods did not fully exploit the benefits from periodic features. Consequently, directly applying them to industrial scenarios would lead to sub-optimal performance.
% poses challenges due to the unique temporal dynamics and periodicity inherent in industrial processes.
\begin{table*}[t]
  \large
  \centering
  \renewcommand\arraystretch{1.2}
  \begin{tabular}{l c r r c c c c}
    \toprule
    \multirow{2}{*}{Dataset}& \multirow{2}{*}{Year} & \multirow{2}{*}{Frames} &\multirow{2}{*}{Resolution} &\multirow{2}{*}{Content} &\multirow{2}{*}{Data Type} &Anomaly&Periodic\\
     & &  &  &  &&classes& information\\
    \hline
    Subway~\cite{Subway} & 2008 & 125,475 & $512\times384$ & Pedestrians& Video  &8& \ding{55} \\ 
    UCSD Ped~\cite{UCSDped} & 2010 & 18,560 & $360\times240$ & Campus & Video &10& \ding{55} \\
    CUHK Avenue~\cite{CUHK} & 2013 & 30,652 & $640\times360$ & Pedestrians & Video &5& \ding{55} \\
    ShanghaiTech~\cite{ShangHaiTech} & 2017 & 317,398 & $856\times480$ & Campus & Video&11&  \ding{55} \\
    MVTec AD~\cite{bergmann2019mvtec} & 2019 & 5,354 & multiple & \textbf{Industrial} & \textbf{Image}&-&  \ding{55} \\
    Street Scene~\cite{Street} & 2020 & 203,257 & $1280\times720$ & Traffic & Video&17&  \ding{55} \\
    ITTB Corridor~\cite{IITB} & 2020 & 483,566 & $1920\times1080$ & Pedestrians & Video&10&  \ding{55} \\
    BTAD~\cite{mishra2021btad} & 2021 & 2,830 & multiple & \textbf{Industrial} &\textbf{Image} &-& \ding{55} \\
    UBnormal~\cite{UBnormal} & 2022 & 236,902 & $1280\times720$ & Human action & Video&22& \ding{55}  \\
    NWPU Campus~\cite{NWPU} & 2023 & 1,466,073 & multiple & Campus& Video &28& \ding{55}  \\
    Real-IAD~\cite{Real-IAD} & 2024 & 151,050 & multiple & \textbf{Industrial}& \textbf{Image} &30& \ding{55}  \\
    IPAD~(Ours) & 2024 & 597,979 & \textbf{$2492\times988$} & \textbf{Industrial} & \textbf{Video}&\textbf{39}& \Checkmark  \\
    \bottomrule
  \end{tabular}
  \caption{Comparison with existing datasets. To the best of our knowledge, we are the first to propose a video anomaly detection dataset for industrial scenarios. Periodicity information is also introduced to adapt to real factory scenarios.}
  \label{tab:dataset_compare}
  \vspace{-6mm}
\end{table*}

To address these challenges and advance the role of VAD in enhancing the manufacturing industry, we first introduce an industrial-specific video anomaly detection dataset. The dataset contains both synthetic and real-world data, including 16 distinct devices commonly found in industrial settings, each accompanied by corresponding anomalies specific to the respective devices. Some of the
normal and abnormal cases in our dataset are shown in Figure~\ref{fig:dataset show}. To ensure the usefulness of the dataset, all devices and anomalies are representative of actual production environments, chosen through on-site factory research and discussions with engineers. 
To account for these real-world challenges, our dataset includes variations in illumination and camera jitter as part of the training and testing sets. We deliberately define these variations as normal conditions to help VAD algorithms distinguish between genuine anomalies and common environmental factors that may induce changes in video appearance. In addition, we field collected four sets of real data that have correspondence with the synthetic data. In this way, we hope to verify the value proposed by the synthetic data.

To address the misfit of existing methods in industrial scenarios, we propose a novel video anomaly detection framework that investigates the significance of periodic features. By introducing a periodic memory module and a sliding window inspection mechanism into the reconstruction-based VAD model, we implicitly and explicitly explore the periodic information to accomplish the VAD task. Through extensive experimentation on our industrial video anomaly detection dataset, we demonstrate that our proposed method outperforms existing VAD models. Meanwhile, we explore parameter-efficient fine-tuning to quickly migrate models trained on synthetic data to real scenarios, which can validate the feasibility and usefulness of our synthetic data in real-world scenarios.

% Finally, we summarize and highlight the key points of the task of this paper. \textcolor{blue}{1)} Our primary objective is to introduce a detection model designed for straightforward training in diverse industrial scenarios, specifically tailored for the OCC task in VAD. \textcolor{blue}{2)} The synthetic dataset we propose aims to evaluate the model's performance by simulating periodic actions of industrial equipment, rather than replicating an actual factory environment. \textcolor{blue}{3)} We emphasize that achieving uniformly well-trained model parameters is not mandatory for this task. Instead, the focal point is a highly adaptable framework that can be flexibly applied to various scenarios.

The main contributions of this paper include:
% \vspace{-4mm}
\begin{itemize}
  \item[\textbullet] We propose a dataset with real and synthetic data for video anomaly detection in industrial scenarios. To the best of our knowledge, this is the first video anomaly dataset for industrial scenes.
  % \item We propose new additional information --- periodicity information, for video anomaly detection algorithms in industrial scenarios, which is reflected in the labeling of the dataset.
  \item[\textbullet] We propose a novel framework that marries the traditional reconstruction-based model with periodic information by introducing a period memory module and a sliding window inspection mechanism. 
  % A transformer adapter is also introduced to speed up the hybrid training for real-scene applications.
  \item[\textbullet] Our work serves to fill the existing gap for video anomaly detection in industrial settings. Utilizing the synthetic data, we explore an efficient finetuning approach to swiftly migrating models to real-world scenarios. 
\end{itemize}

\section{Related Work}
\label{sec:related}

\subsection{Video Anomaly Detection}

In the field of video anomaly detection, researchers typically leverage video datasets containing both normal and annotated anomaly videos for training anomaly detection models. The availability of rich and detailed labeling in the anomaly data often contributes to the models exhibiting good performance in the test set~\cite{sultani2018real,zaigham2021cleaning,zaheer2020claws,zhong2019graph}. Due to the challenge of capturing anomalies in various scenarios, researchers have introduced datasets that exclusively contain normal videos in the training set, aiming to reduce the reliance on labeled abnormal data, which is called One-Class Classification (OCC) task. Existing widely used video datasets~\cite{Subway,UMN,UCSDped,CUHK,ShangHaiTech,Street,IITB,UBnormal,NWPU} primarily focus on pedestrian and traffic scenarios, with a significant emphasis on human-centric anomalies. Therefore, it lacks a complete video anomaly detection dataset for industrial scenarios.

The mainstream frameworks for OCC problem can be broadly categorized into two classes: reconstruction-based models and prediction-based models. Autoencoder~(AE) structure~\cite{AE} was firstly used for video frame reconstruction in the context of anomaly detection. MemAE~\cite{MemAE} introduced a memory structure to enhance the optimization of feature information in the reconstruction process. In addition to AE, some methods utilize Generative Adversarial Networks (GANs)~\cite{GAN} for video frame reconstruction, enabling differentiation of anomalies based on reconstruction results~\cite{SVD-GAN,liu2018future,lu2020few}. To recognize the object motion more accurately, optical flow is often incorporated to enhance recognition of object motion and guide detection of anomalous movements~\cite{liu2021hybrid,li2021variational,chang2020clustering,zou2021progressive}. Recently, 
% there has been a growing consideration for scene dependency in VAD. S
some methods focus on extracting background information separately to analyze anomalies within different scenes~\cite{NWPU, sun2023hierarchical}. However, due to the lack of periodicity information, all these methods may encounter challenges when dealing with industrial scenarios that present strong periodic characteristics.

\subsection{Image-Based Industrial Anomaly Detection}
% Anomaly detection plays a crucial role in industrial production processes, aiming to detect and identify anomalies that deviate from normal operating conditions in a timely manner. 
In recent years, data-driven methods for industrial anomaly detection have gained significant attention. There are several image-based datasets~\cite{bergmann2019mvtec,mishra2021btad} for industrial anomaly detection. These datasets primarily focus on single-image anomaly detection, where the goal is to jointly determine the presence of an anomaly and localize the specific regions. This line of work is known as anomaly localization~\cite{tao2022deep}.
The dominant approaches to this task can be generally categorized into four branches: image reconstruction-based model~\cite{tan2021trustmae,wang2020image,liu2021unsupervised,tsai2022multi,Anomaly-Segmentation}, image generation-based model~\cite{liu2020towards,schlegl2017unsupervised,yu2021fastflow,gudovskiy2022cflow}, deep feature embedding based model~\cite{bergmann2020uninformed,yan2021unsupervised,rippel2021transfer,kim2021semi,salehi2021multiresolution} and self-supervision learning based model~\cite{li2021cutpaste,pirnay2022inpainting}. 
% The aforementioned anomaly detection methods have demonstrated remarkable performance by achieving detection accuracy rates exceeding 99$\%$ on the MVTec AD dataset. 
However, these methods primarily focus on detecting appearance anomalies within still images and do not explicitly incorporate temporal information from videos. Therefore, it is unfeasible to directly apply these methods to video anomaly detection in industrial scenarios.

\begin{figure}[t]	
    \centering	
	\includegraphics[width=1.0\linewidth]{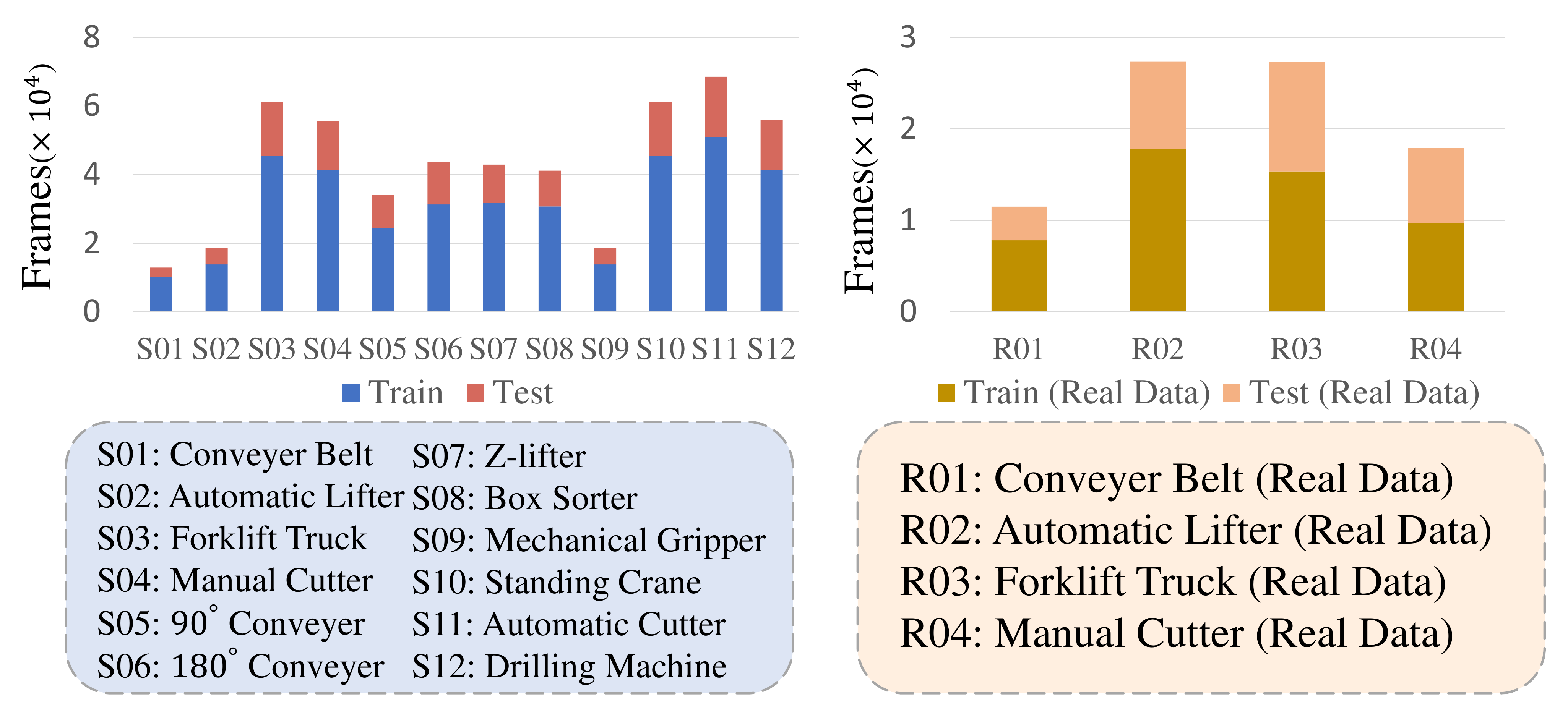}
    \vspace{-4mm}
	\caption{Dataset statistics. The bar chart shows the percentage of train data and test data in different cases. The specific devices in each case are given in the table.}
    \label{fig:dataset statistics}
    \vspace{-4mm}
\end{figure}

\subsection{Synthetic Data in Video Understanding}
% There are many advantages of synthetic data over real data. In addition to the ease of data collection, we can generate synthetic data that is difficult to capture in the real world and occurs infrequently in nature but is critical for training. As a result, 
Many video datasets constructed from synthetic data are emerging in the field of video understanding, due to the flexibility of generating diverse and large amounts of videos. Virtual KITTI~\cite{gaidon2016virtual} is a widely used video synthesis dataset well suited for many video understanding tasks. CARLA~\cite{dosovitskiy2017carla} proposed an open-world simulator that can capture synthetic video data by customizing the position and viewpoint of the camera.  In the video anomaly detection task, UBnormal~\cite{UBnormal} is an excellent work based on a synthesized dataset. It emulates various action behaviors, including those of pedestrians and vehicles, and establishes corresponding abnormal events.

These publicly accessible synthetic datasets exemplify the viability of utilizing simulated data for video understanding tasks. Consequently, we also construct a synthetic dataset, aiming to overcome the challenges associated with data acquisition in industrial scenarios. Nevertheless, it is important to acknowledge the limitations of synthetic data, particularly in terms of disparities with real-world data. To mitigate the effects of such discrepancies, we have incorporated real-scene video data into our dataset as well.

\section{Dataset}
\subsection{Description of Dataset}
The comprehensive dataset comprises a blend of real and virtual data. The real data was acquired through live photography collection, while the virtual data was meticulously crafted using SolidWorks software, resulting in the creation of 16 distinct models of industrial equipment.
% The dataset construction process involved utilizing SolidWorks\footnote{www.solidworks.com} software to create 12 distinct models of industrial equipment (from VAD01 to VAD12). 
These models encompass a range of machinery such as conveyor belts, lift tables, cutting machines, drilling machines, machine grippers, cranes, \emph{etc}. The animation simulation feature within the SolidWorks software enabled us to design specific cycle actions and abnormalities for each device. By setting the desired action trajectories, we were able to generate videos by rendering the 3D models. In contrast to capturing live video data from factories, the animation generation process in SolidWorks offers greater flexibility and control. To replicate real-world conditions, we utilized controlled light source points. This enabled us to simulate realistic changes in light and variations in viewpoints. To bridge the gap between the simulated and real-world data, we incorporated four realistic cases captured by a camera (from R01 to R04), resulting in a total of 16 different devices. These real data scenarios all have counterparts in the synthetic data (from S01 to S04), which are used to explore the migration effect of the model under different data dimensions.
% Indeed, the security and privacy concerns associated with acquiring real-world data in factory settings pose significant challenges. The collection and public dissemination of such data become more difficult due to these constraints. Considering these factors, we strongly believe that animation simulation is a more practical and viable alternative for data collection.

To align with the distribution of anomalous data in real-world scenarios, we only include normal cases in the training set. In the test set, we intentionally introduced a wide range of anomalies to evaluate the performance of the anomaly detection system: appearance anomalies (\eg, color change, shape change), position anomalies (\eg, position shift, angle tilt), short-term motion anomalies (\eg, lagging, a large change in speed), and logic anomalies (\eg, sorting error, wrong sequence of actions). Additionally, we designed specific anomalies for different types of industrial equipment in our dataset. For example, the cutting machine cuts in the wrong position, the forklift truck does not lift the object successfully. In addition to the previously mentioned anomalies, we have also incorporated lighting variations and viewpoint jitter as normal cases in both the training and testing sets. These variations introduce significant pixel-level variations in the video frames. Some of the normal and abnormal cases in our dataset are shown in Figure~\ref{fig:dataset show}.

\subsection{Dataset Statistics}

% Table.~\ref{tab:dataset_statistics} shows the distribution of the number of video frames in our dataset. 
IPAD dataset comprises a total of 597,979 frames, with 430,867 frames allocated for training data and 167,112 frames for the test data. The quantitative comparison between our datasets and other VAD datasets is reported in Table~\ref{tab:dataset_compare}. Compared to most existing datasets, our dataset boasts a larger data size, surpassing the majority of them, with the exception of NWPU Campus. Additionally, we excel in terms of video resolution and anomaly classes, outperforming all other datasets in this aspect. For the main content of the video, to the best of our knowledge, we are the first to propose a video anomaly detection dataset that focuses on industrial scenarios. Figure~\ref{fig:dataset statistics} presents the distribution of data across different industrial equipment scenarios within the dataset, which includes a total of 16 different devices (12 animated models and 4 real data scenario). During collection, we ensure that the data is relatively evenly distributed among the various scenarios. A notable difference between our dataset and existing multi-scene datasets is that the primary content of the frames remains relatively consistent across different scenes in existing datasets. However, there are significant variations between scenes in IPAD. 

\begin{figure*}[t]	
    \centering	
	\includegraphics[width=1.0\linewidth]{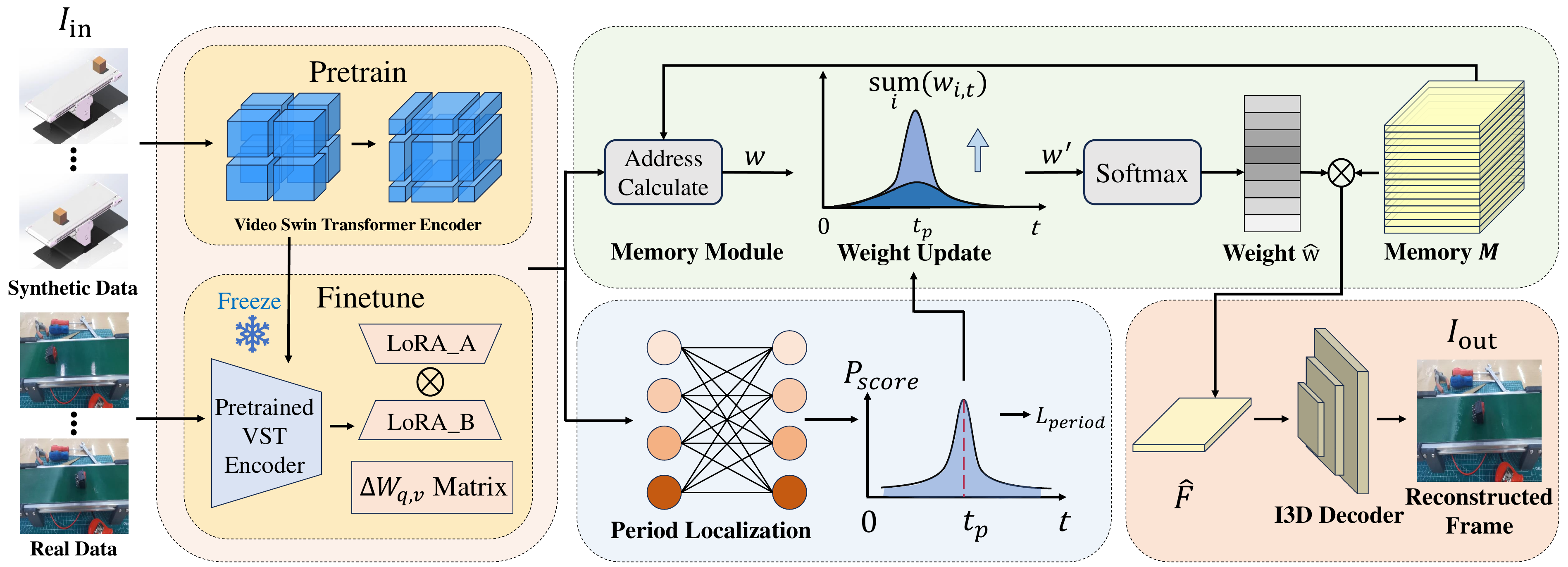}
    \vspace{-6mm}
	\caption{The proposed reconstruction-based model with periodic information. The model undergoes an initial pre-training phase using synthetic data. Subsequently, the encoder segment is fine-tuned through a combination of real data and LoRA~\cite{hu2021lora}, enhancing its performance and adaptability. The features of the input video clip are first extracted by the video swin transformer to get $F$. Then addressing to the memory module and period classification are performed separately to get the initial weights $W$ and period relative position ${t}_{p}$. Then the weights are updated to get $\hat{w}$, which in turn is computed to get the output feature $\hat{F}$. Finally, the reconstructed video frame is obtained using the I3D decoder.}
    \label{fig:pipeline}
    % \vspace{-2mm}
\end{figure*}

Regarding data labeling, we initially labeled the test data at the frame level to facilitate the evaluation of the model performance. Additionally, since our video data exhibits a pronounced cyclical nature, we conducted periodic cropping on the videos within both the training and test sets. The operation cycles of different devices vary in length, ranging from 5 seconds to 30 seconds. However, it is important to note that the cycle time length remains relatively fixed within each specific device, except for cases where anomalies occur, leading to variations in the cycle length. Certainly, periodicity is a noteworthy timing feature in industrial scenarios, and providing relevant annotations to indicate the periodic nature of the data can greatly contribute to improving the performance of anomaly detection methods.
%Although we also performed cycle cropping on the test set, it is clear that in practice it is almost impossible to preprocess unknown test data for cropping. Therefore, we do not encourage algorithmic models applied to this dataset to use cycle labeling on the test set.

Overall, our proposed dataset has several advantages over other existing video anomaly detection datasets: 1) new video content (industrial); 2) both real and synthetic data; and 3)
  % \item Lighting changes and viewpoint jitter.
rich periodic data and corresponding labels.
% \end{enumerate}

\begin{figure}[t]	
    \centering	
	\includegraphics[width=1.0\linewidth]{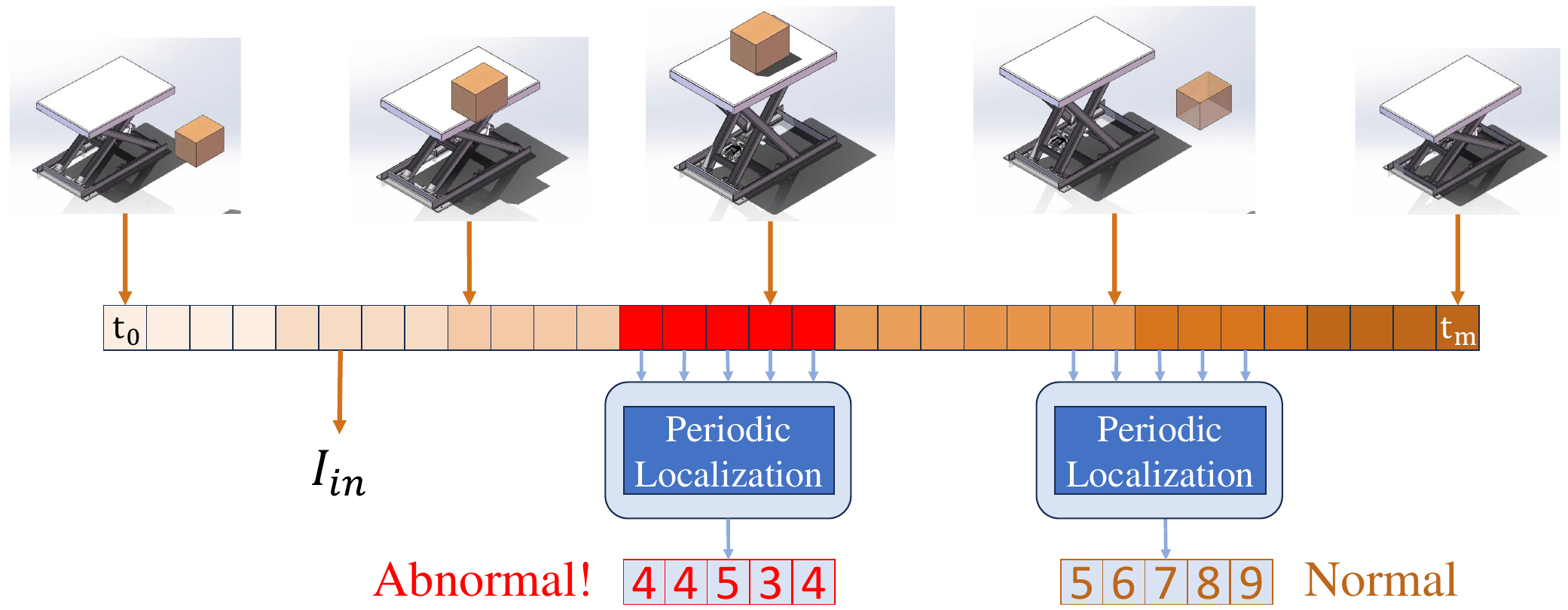}
    \vspace{-4mm}
	\caption{Sliding window. The sequence represents an action cycle, where each square represents an input video clip.}
    \label{fig:slide window}
    \vspace{-6mm}
\end{figure}

\section{Method}
Existing methods in the field of analyzing human-centric data are typically designed to handle irregular patterns and do not explicitly incorporate periodic information. However, in the operation of industrial equipment, cyclic movements play a central role, offering a valuable opportunity to leverage long-periodical information for anomaly detection. Hence, we hypothesize that exploiting the periodic feature can assist in localizing short video sequences within the corresponding positions of the period. 
% To this end, we first introduce reconstruction-based structure to achieve the task basically in Section~\ref{sec:reconstruction}. In Section~\ref{sec:memory}, we present a periodic memory module to save and combine period information to reconstruction model implicitly. To utilize periodic information explicitly, in Section~\ref{sec:window}, we propose a sliding window inspection to detect the periodicity of the test video sequence, and thus increase the accuracy. Finally, in Section~\ref{sec:Adapter}, to make the model more efficient in accomplishing the migration of data dimensions, we introduce a transformer adapter in the encoder structure.

\subsection{Reconstruction Model}
\label{sec:reconstruction}

Our model follows a reconstruction-based structure, leveraging the Auto-Encoder (AE) as its fundamental architecture. The input video sequence is initially processed by the encoder to extract spatio-temporal features, resulting in a high-dimensional feature vector. This feature is subsequently utilized as input for image reconstruction, accomplished by the decoder component. The reconstructed image is then compared with the original input video frame, enabling the computation of an anomaly score. 
In the feature extraction stage, conventional methods typically employ a 3D convolution (or 2D convolution for single images) to construct the encoder. However, in our approach, we opt to utilize the Video Swin Transformer as the feature extractor, inspired by its exceptional feature extraction capabilities demonstrated in~\cite{video-swin-T}. The Transformer structure enables effective feature extraction by leveraging the inherent strengths of self-attention mechanisms. For the decoder part, we use the inverse convolution structure of I3D~\cite{I3D}.

\begin{table*}[t]
  \centering
  \renewcommand\arraystretch{1.2}
  \begin{tabular}{l c c c c c c c c c c c c c c c c c c c}
    \toprule
     % Method&  Device01 & Device02 & Device03 & Device04 &  Device05 & Device06 & Device07 & Device08 & Device09 & Device10 &  Device11 & Device12 & Device13 & Average\\
     Method&  S01 & S02 & S03 & S04 &  S05 & S06 & S07 & S08 & S09 & S10 &  S11 & S12 & R01 & R02 & R03 & R04 & Avg.\\
    \hline
    conAE~\cite{AE} &  63.1 & 47.7 & 53.0 & 34.7 & 82.9 & 46.6 & 58.3&  69.1 & 53.0 & 55.3 & 39.8 & 50.4 & 77.6 & 64.3 & 40.7 & 70.1 & 56.7 \\
    memAE~\cite{MemAE} &  63.2 & 50.6 & 65.6 & 49.5 & 78.8  & 45.9 & 57.9&  84.7 & 65.7 & 59.9 & 49.4 & 50.7 & 77.9 & 65.0 & 41.6 & 70.7 & 61.1 \\
    AstNet~\cite{ASTNet} &  67.7 & 52.0 & 61.0 & 51.6 & 80.4 & 54.1 & 54.5&  82.6 & 59.8 & 55.7 & 47.8 & 60.8 & 79.8 & 66.8 & 42.1 & 67.6 & 61.5 \\
    DMAD~\cite{DMAD}  &  55.9 & 55.3 & 47.9 & 47.9 & 69.3 & \underline{61.0} & \textbf{66.9}&  \underline{87.5} & 69.7 & \textbf{67.0} & \underline{56.0} & 55.8 & 79.5 & 68.5 & \underline{43.1} & 63.1 & 62.2\\
    V-Swin-T~\cite{video-swin-T} &  \underline{68.2} & \underline{60.0} & \underline{66.6} & \underline{54.7} & \underline{85.6} & 53.3 & 59.5&  \textbf{88.5} & \underline{69.7} & 60.5 & 54.8 & \textbf{69.1} & \underline{81.1} &\underline{74.1} & 42.3 & \underline{75.5} & \underline{66.5}  \\
    Ours &  \textbf{69.5} & \textbf{63.9} & \textbf{70.6} & \textbf{58.3} & \textbf{86.2} & \textbf{61.2} & \underline{60.6}&  85.6 & \textbf{71.2} & \underline{62.2} & \textbf{60.9} & \underline{67.1} & \textbf{84.4} & \textbf{75.4} & \textbf{43.5} & \textbf{76.7} & \textbf{68.6} \\
    \bottomrule
  \end{tabular}
  \caption{Main Result on the industrial period video dataset. Each result is frame-level AUC score. Best results are highlighted in bold and the second-best result is underlined. The last column indicates the mean value of AUC under all cases.}
  \label{tab:main result}
  \vspace{-6mm}
\end{table*}

\subsection{Periodic Memory Structure}
\label{sec:memory}

To leverage the periodic features of video sequences, we introduce a periodic memory structure into our reconstruction framework. The specific architecture of this memory module is illustrated in Figure~\ref{fig:pipeline}. In real-world scenarios, the length of the action period of industrial equipment cannot be determined. Therefore we need to design a generalized structure to extract and utilize the periodic information. To address this challenge, we draw inspiration from the mem-AE approach~\cite{MemAE} and adopt a strategy of saving and retrieving the features of the complete cycle in temporal order during the training phase. we initially perform video clip localization in terms of temporal periods and associate the corresponding memory addresses with features extracted by the encoder. Subsequently, we update the weights of the memory addresses based on the acquired period localization results. These updated addresses are then utilized to effectively select features stored in the memory.

Specifically, to align the feature $F_{in}$ obtained after feature extraction by the video swin transformer with the size of the memory module, we perform a dimensional transformation, resulting in $\hat{F}_{in}$ with dimensions $\mathbb{R}^{T\times C}$. Additionally, we define the memory module as $\hat{M}$ with dimensions $\mathbb{R}^{M\times C}$, which stores $M$ features. During the addressing process, we calculate the preliminary feature weight $w$:
\begin{equation}
  w = \hat{F}_{in} \times \hat{M}^T, w\in \mathbb{R}^{T\times M}.
  \label{eq:weight1}
\end{equation}

In the period classification step $f_{p}(\cdot)$, we aim to determine the relative position of an input video segment within a cyclic motion. Since the length of each cycle may vary, we need to map it to a fixed range to ensure consistent comparisons. To minimize the accuracy degradation caused by setting too many ranges, this mapping is typically a surjective rather than an injective function. Consequently, we obtain the relative position $t_p$ of the video segment within the cycle, taking into account the cyclic nature of the motion and enabling us to analyze its temporal context accurately:
\begin{equation}
  t_p = \text{argmax}_{P_{s}} f_{p}(F), t_p\in (0,t_{\max}),
  \label{eq:tp1}
\end{equation}
where $P_s$ denotes the period localization score.
After that, map $t_p$ under the memory module size range, \ie, $f_t(\cdot):t_{max}\to M$. Based on the obtained relative positions of the cycles, the weights of $w$ are updated to obtain $w^{'}$:
\begin{equation}
  {t}^{'}_p = f_t(t_p), {t}^{'}_p\in (0,M),
  \label{eq:tp2}
\end{equation}
for all $i\in [0,T]$:
\begin{equation}
  {w}^{'}_{i,{t}^{'}}=\left\{
    \begin{aligned}
    &{w}_{i,{t}^{'}}&&{if\,{t}^{'}\ne {t}^{'}_p}\\
    &{w}_{i,{t}^{'}}\cdot (1+P_{s,t_p})&&{if\,{t}^{'}={t}^{'}_p}\\
    \end{aligned}
    \right.,
  \label{eq:weight2}
\end{equation}
where ${w}^{'}_{i,{t}^{'}}$ represents the element located in row $i$ and column ${t}^{'}$ of the ${w}^{'}$ matrix. By examining Equation~\ref{eq:weight2}, it becomes evident that the memory weights are primarily updated in the column vector dimension. To obtain the final weights $\hat{w}$, we apply the softmax function to the column vector of ${w}^{'}$. This operation is performed for all $j$ values within the range of $[0,M]$. So $\hat{w}$ can be specified as:
\begin{equation}
  \hat{w}_{i,j} = \frac{\exp({w}^{'}_{i,j})}{\sum_{i=0}^{T} \exp({w}^{'}_{i,j})}.
  \label{eq:weight3}
\end{equation}

The feature vectors $\hat{F}_{out}$ corresponding to the cycles in the memory module are extracted based on the weights $\hat{w}$. We calculate $\hat{F}_{out}$ by multiplying the features stored in the memory module memory module $\hat{M}$ with the weight $\hat{w}$:
\begin{equation}
  \hat{F}_{out} = \hat{w}\times \hat{M}, \hat{F}_{out}\in \mathbb{R}^{T\times C}.
  \label{eq:Fout}
\end{equation}

To complete the reconstruction of the target frame, the obtained feature vector $\hat{F}_{out}$ is passed through the decoder part, which is structured similarly to the I3D (Inflated 3D) architecture~\cite{I3D}. The decoder takes the transformed feature vector $\hat{F}_{out}$ as input and leverages its information to reconstruct the target frame. It is important to note that the `$\times$' symbol used in the previous formulas denotes matrix multiplication.

\subsection{Sliding Window Inspection}
\label{sec:window}

After obtaining the relative positions of the input frames within the cycle using the method described in Section~\ref{sec:memory}, we employ a sliding window inspection approach to detect the periodicity of the test video sequence. In addition to implicitly introducing periodic information at the feature level, we try to utilize periodic information in a more explicit way. The specific structure of the sliding window is illustrated in Figure~\ref{fig:slide window}. Following the cycle classification process described in Section~\ref{sec:memory}, we obtain the prediction of the relative position of the cycle for each input video segment, denoted as ${t}_{p}$. When considering an input target frame, which is essentially a short video clip, we can obtain an array of relative positions of cycles centered around it. This array is denoted as ${T}_{p}^{nt}$, where $n$ represents the window size, and $t$ denotes the $t$-th target frame. Typically, an odd value is chosen for $n$ to ensure a symmetric window around the target frame. So ${T}_{p}^{nt}$ can be specified as:
\begin{equation}
  {T}_{p}^{nt} = [t_{p}^{t-\frac{n-1}{2}}, t_{p}^{t+1-\frac{n-1}{2}},\cdots,t_{p}^{t+\frac{n-1}{2}}].
  \label{eq:Tpnt}
\end{equation}

Obviously, when the input video segment is the normal case, we will theoretically get an increasing isotropic series. On the contrary, when it is abnormal, the relative positions of its cycles will change irregularly. Therefore, in order to show this difference, we calculate the difference between the period prediction series and the normal period series:
\begin{equation}
  E_{p} = \frac{1}{n}\sum\nolimits_{i=-\frac{n-1}{2}}^{\frac{n-1}{2}} \lvert t_{p}^{t+i}-{b}^{t+i}\rvert,
  \label{eq:window error}
\end{equation}

where $E_p$ denotes the period localizing sequence error, $b$ refers to the pre-defined temporal reference sequence. When the length of the action cycle in the video is approximately equal to $t_{\max}$, $b$ is defined as $[t_{p}-(n-1)/2,t_{p}+1-(n-1)/2,\cdots,t_{p}+(n-1)/2]$. This sequence provides a reference for analyzing the relative positions of cycles within the video.

\begin{table*}[t]
  \centering
  \large
  \begin{tabular}{c c c c c c c c c c c}
    \toprule
      \multirow{2}*{Pretrain Data} & \multicolumn{2}{c}{Finetune Data} & \multirow{2}*{Finetune Method} & \multirow{2}*{R01} & \multirow{2}*{R02} & \multirow{2}*{R03} & \multirow{2}*{R04} & \multirow{2}*{Avg.} & Finetune & Trainable\\
      \cline{2-3}
      % \cline{10-11}
      & Synthetic & Real &&&&&&& Time$\downarrow$ & Parameters$\downarrow$\\
    \hline
    \ding{55} & \ding{55} & All & \ding{55} & 84.4 & 75.4 & 43.5 & 76.7 & 70.0 & 17min & 35.9M\\
    \ding{55} & S01-S04 & \ding{55} & \ding{55} & 30.5 & 62.1 & 40.5 & 66.8 & 49.9 & 37min & 35.9M\\
    \ding{55} & S01-S04 & All & \ding{55} & 83.7 & 75.1 & 43.8 & 75.4 & 69.5  & 53min & 35.9M\\ \hline \hline
    S01-S04 & \ding{55} & All &Fully Finetune & 87.9 & 75.9 & 44.1 & 77.1 & 71.3 & 17min & 35.9M\\
    S01-S04 & \ding{55} & Few Shot& Fully Finetune & 85.1 & 72.6 & 42.8 & 72.2 & 68.2 & 4min & 35.9M\\
    S01-S04 & \ding{55} & Few Shot & PEFT & 86.3 & 69.4 & 42.6 & 70.4 & 67.2 & 2.3min & 10.3M\\
    {\{S01-S04\}}$^{*}$ &\ding{55} & Few Shot & PEFT & 85.4 & 69.9 & 44.5 & 70.6 & 67.6  & 2.3min & 10.3M\\
    {\{S01-S12\}}$^{*}$ &\ding{55} & Few Shot & PEFT & 85.7 & 70.7 & 46.0 & 71.1 & 68.4  & 2.3min & 10.3M\\
    
    % unreal train & 30.5 & & 45.5 & 66.8\\
    % DA & & & &\\
    % Peft
    \bottomrule
  \end{tabular}
  \caption{ The result of hybrid learning. PEFT means Parameter-Efficient Fine-Tuning. $^{*}$: {\{\}} denotes the training of a unified pretrained model using the entirety of the data available. Others indicate that each case involves individual models.}
  \label{tab:Migration learning}
  \vspace{-6mm}
\end{table*}

% During anomaly detection, traditional reconstruction methods may struggle to identify certain situations, such as lighting changes. These changes can result in substantial pixel variations in the video frames, leading to higher reconstruction errors and consequently higher anomaly scores. Indeed, when considering the periodic nature of actions, changes such as lighting variations may not significantly disturb the primary periodic features of the action. While traditional reconstruction methods may struggle to handle these changes, leveraging periodic information can greatly enhance the performance.

\subsection{ Adapter for Parameter-efficient Finetuning}
\label{sec:Adapter}

To enhance the model's anomaly detection accuracy in real-world scenarios, we opted for a training approach involving pretraining with synthetic data followed by finetuning with few-shot real data. However, integrating a large influx of data can exert significant training pressure on the model. To streamline the finetuning process with real data, we introduce an adapter structure suitable for transformers in the encoder segment. Drawing inspiration from LoRA~\cite{hu2021lora}, we initially freeze most pretraining parameters of the encoder, retaining only the Embedding and LayerNorm components. Within the attention module, we treat $q$ and $v$ as low-rank matrices, utilizing the product of two low-dimensional matrices as their training parameters. This strategy allowed us to substantially minimize training parameters while maximizing model accuracy.

\section{Experiments}

\subsection{Dataset and Evaluation Protocol}
We perform experiments on the proposed IPAD dataset. During the test phase, we utilize the area under the curve (AUC) to evaluate the performance of VAD methods. AUC is a widely adopted evaluation metric commonly used to assess the performance of binary classification models. It quantifies the area under the receiver operating characteristic (ROC) curve generated by the model. We concatenate the video frames in the test set and calculate the AUC values. It is important to note that the videos in the test set are divided into multiple shorter video segments. In many existing datasets, each short video segment contains a mix of normal and abnormal frames. Consequently, the reconstruction differences within these short video segments can vary significantly, so only this video segment needs to be normalized when calculating the anomaly score. Indeed, considering the characteristics of our dataset where there are more all-abnormal or all-normal video segments, normalizing the short segment may not provide meaningful results.
%\textcolor{red}{it may not be appropriate to calculate the anomaly score by normalizing an entire segment of normal video. In cases where the entire video segment contains only abnormal frames or only normal frames, normalizing the entire segment may not provide meaningful results. This is because normalization assumes a mixture of normal and abnormal frames within the segment, and normalizing an all-normal segment could lead to misleading anomaly scores.}
To align with the specific characteristics of our dataset, we have decided to normalize the reconstruction error results for the entire test set within the same scenario. 
%This approach is more reasonable compared to normalizing the reconstruction error for short video segments.

\begin{figure*}[t]	
    \centering	
	\includegraphics[width=1.0\linewidth]{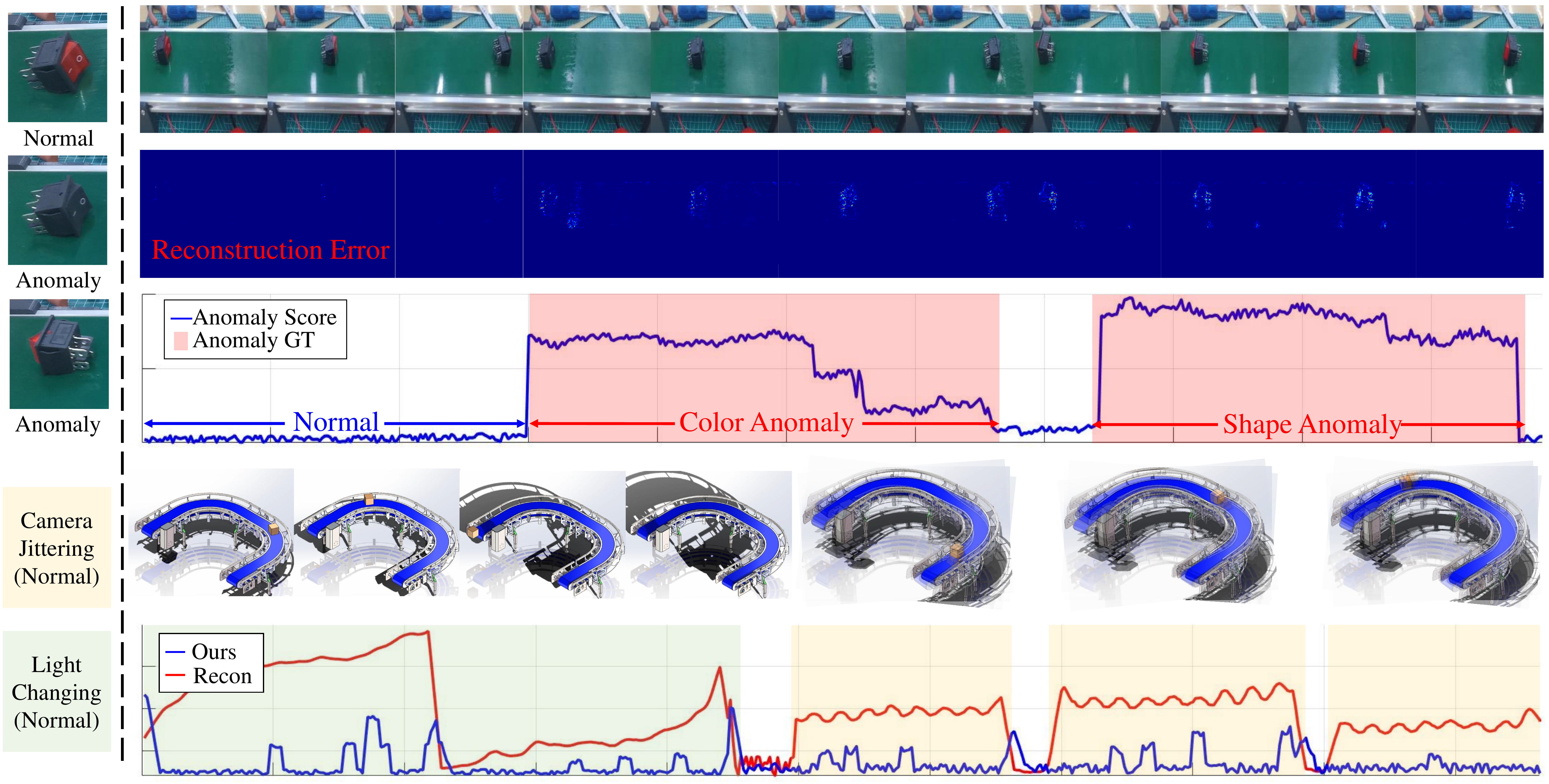}
    \vspace{-4mm}
	\caption{Anomaly score curves. The first row shows the detection results of two anomalies. The second row shows the case of light change and camera jitter, and it can be seen that our method has lower anomaly scores, which is closer to GT.}
    \label{fig:light}
    \vspace{-2mm}
\end{figure*}

\subsection{Implement Details}
Our dataset comprises diverse content captured from different devices, which results in significant variations in the data characteristics. As a result, training and testing the same model on the entire dataset may not yield optimal results due to the content variations. Hence, we trained and tested separate models, with each model weight corresponding to a specific scenario.
% For the video segment input, we take 16 frames as the input video clip and resize each frame to $256\times256\times3$. For the encoder part, we use Video Swin-T ($C=96,layer\,numbers=\lbrace2,2,6,2\rbrace$) structure for feature extraction. The decoder section uses the I3D structure for video frame reconstruction. Meanwhile, we use the periodic memory module mentioned in Sec.~\ref{sec:memory}, where the mapping category of cycles is set to 200 and the memory dimension to 2000. For the Sliding Window part mentioned in~\ref{sec:window}, we set the window size to 5. In the training process, we choose batch size as 8 and train 50 epochs for each scene.
For the video segment input, we adopt a configuration where each input video clip consists of 16 frames. We resize each frame to a dimension of $256\times256$.
By default, the encoder part of our model employs the Video Swin-T~\cite{video-swin-T} architecture with specific parameters: $C=96$ and layer numbers $\lbrace2, 2, 6, 2\rbrace$. Furthermore, we incorporate the periodic memory module described in Section~\ref{sec:memory}, with the number of mapping categories set to 200. The memory dimension is set to 2000. In addition, we employ the Sliding Window technique mentioned in Section~\ref{sec:window}. The window size is set to 5. During the training process, we set a batch size of 8 and Adam is employed as the default optimizer with an initial $1e{-4}$ learning rate. The model is trained for 50 epochs for each scene.

% \begin{table}[t]
%   \centering
%   \renewcommand\arraystretch{1.2}
%   \begin{tabular}{c c c c}
%     \toprule
%      Recon.& Period. & Synthetic Data & Real Data\\
%     \hline
%     \ding{55} & \checkmark & 60.4 & 77.8\\
%      \checkmark& \ding{55}& 65.9 & 81.1\\
%     \checkmark & \checkmark & \textbf{68.1}& \textbf{84.4}\\
%     \bottomrule
%   \end{tabular}
%   \caption{Comparison of the AUC results for whether or not to add periodic information to the reconstruction-based model.}
%   \label{tab:ablation result}
% \end{table}

\subsection{Comparison with State-of-the-arts}

The main experimental results are summarized in Table~\ref{tab:main result}, showcasing the performance of our proposed methods. In our experiments, we constructed an auto-encoder reconstruction structure utilizing the Video Swin Transformer as the feature extractor, referred to as `V-Swin-T'. Additionally, we enhanced this structure by incorporating periodic information using the periodic memory module and sliding window inspection, denoted as `Ours'. Many methods in recent years have incorporated human pose information to enhance accuracy on existing datasets~\cite{xiao2023divide}. However, it is evident that our dataset emphasizes industrial devices rather than human behavior. ASTNet~\cite{ASTNet} stands out as the best open-source method without relying on additional information. We complement our experiments with DMAD~\cite{DMAD}. It is important to note that this experiment includes background annotation.

Our approach demonstrates superior performance on the proposed dataset, achieving an average AUC of 68.6$\%$. This signifies the effectiveness and optimality of our proposed methods in accurately detecting anomalies within the dataset. Compared to previous methods, the transformer encoder employed in our approach enables the extraction of superior spatial and temporal features from the video data. Furthermore, in addition to enhancing the backbone architecture, the inclusion of periodic information plays a crucial role in enabling the model to better understand the underlying actions present in the videos. This improved comprehension of video actions contributes to enhanced anomaly detection performance, as evidenced by the higher AUC achieved in our experiments.

\subsection{Hybrid Learning of Synthetic and Real Data}

Considering the difficulties in collecting real-world training data in industrial scenarios, we explore reducing the unattainable data requirements with the help of synthetic data. Table~\ref{tab:Migration learning} shows the experimental results under this migration task. 
In our dataset, synthetic scenes S01-S04 and real-world cases R01-R04 correspond to each other. Initially, we pursued a one-stage training approach, employing synthetic and mixed data as the training set and testing on the corresponding real data. Despite the incorporation of synthetic data in training, this approach resulted in decreased accuracy (49.9\% and 69.5\%), highlighting the substantial differences between the two data types. 

To better adapt the model to real-world scenes, we adopted a two-stage "pretrain+finetune" training strategy. Synthetic data was utilized for pretraining to impart fundamental motion features, while finetuning with real data tailored the model to real-world dimensions, yielding the highest accuracy of 71.3\%. Given the challenges in real data collection, we aimed to minimize the required data scale by using only 1/5 of the real data for finetuning. 
This approach significantly reduced data needs and finetuning time, with a minor accuracy reduction of 2.1\%. To further mitigate model migration costs, we integrated adapter layers into the encoder, halving the finetuning time while maintaining a mere 1.0\% accuracy drop. However, these experiments necessitated a pretrained model for each real-world scenario, hampering rapid deployment. To address this, we leveraged the extensive virtual data, conducting comprehensive pretraining on S01-S04 and S01-S12 scenes, respectively, and then performing fast fine-tuning in real scenarios. This strategy led to accuracy improvements of 0.4\% and 1.2\%, underscoring the significant value of our extensive synthetic data for real-scene migration. Overall, by pretraining the model and incorporating the adapter structure, we successfully reduced data requirements to 1/5 and training time to 1/7, albeit with a slight trade-off of 1.6\% accuracy loss compared to training solely on real data.

\begin{table}[t]
  \centering
  \renewcommand\arraystretch{1.2}
  \begin{tabular}{c c c c}
    \toprule
     Recon.& Period. & Synthetic Data & Real Data\\
    \hline
    \ding{55} & \checkmark & 60.4 & 63.4\\
     \checkmark& \ding{55}& 65.9 & 66.7\\
    \checkmark & \checkmark & \textbf{68.1}& \textbf{70.0}\\
    \hline
    Window.& Memory. &  & \\
    \hline
    \ding{55} & \checkmark & 65.0 & 66.9\\
     \checkmark& \ding{55}& 66.8 & 68.3\\
    \checkmark & \checkmark & \textbf{68.1}& \textbf{70.0}\\
    \bottomrule
  \end{tabular}
  \captionof{table}{Ablation results for period information block (sliding window and memory module).}
  \label{tab:ablation result}
  \vspace{-10mm}
\end{table}

\subsection{Ablation Study}

To assess the performance enhancement achieved by introducing periodic information, we conducted ablation experiments on both the periodic memory module and sliding windows. The results of these experiments are presented in the upper part of Table~\ref{tab:ablation result}.
When the periodic information is removed, the performance of the model experiences a dramatic drop, resulting in an AUC of 65.9$\%$ and 66.7$\%$. This demonstrates the crucial role played by the periodic memory module in capturing and leveraging periodic information. Besides, The results of the ablation experiments for each of these two methods are given in the lower part of Table~\ref{tab:ablation result}, and the absence of each component brings about a significant decrease in AUC. When the periodic information is reintroduced, the performance is notably improved, resulting in an AUC of 68.1$\%$ and 70.0$\%$. 

In Figure~\ref{fig:light}, the visualization results vividly illustrate the advantages of integrating periodic information into our model. Traditional reconstruction methods often struggle to identify anomalies in scenarios like lighting changes, which can lead to significant pixel variations in video frames, resulting in heightened reconstruction errors and subsequently elevated anomaly scores. Incorporating periodic information equips the model with a deeper understanding of temporal dynamics and actions within video sequences. Notably, the visualization results showcase enhanced stability, particularly in handling lighting variations and viewpoint jitter scenarios. Through the utilization of periodic information, the model attains a more profound grasp of underlying video actions, effectively mitigating these fluctuations.
%Methods based entirely on reconstruction can lead to false detection problems when the effect of illumination becomes progressively larger and the reconstruction error also increases progressively. Methods that incorporate periodicity information can effectively reduce the reconstruction error caused by illumination.
% \begin{figure}[t]	
%     \centering	
% 	\includegraphics[width=1.0\linewidth]{Image/window_size_v2.pdf}
%     \vspace{-6mm}
% 	\caption{AUC results for different window sizes.}
%     \label{fig:window_size}
%     \vspace{-2mm}
% \end{figure}

% In addition to the ablations on memory modules, we also performed a series of experiments to evaluate the impact of window sizes. The experimental results, as presented in Figure~\ref{fig:window_size}, indicate that a smaller window size leads to incomplete period judgments, while a larger size results in a higher level of judgment. As a result, setting the window size to 5 yields the highest AUC. 
\vspace{-2mm}
\section{Conclusion}
%In this paper, to solve the VAD problem in industrial scenarios, we propose an industrial periodic video anomaly detection dataset, which is the first dataset considering industrial scenarios under the VAD task. Furthermore, aiming at the significant periodic characteristics of industrial equipment actions, we propose a method that combines the periodic information with reconstruction-based model. We design the period memory module and sliding window inspection to utilize periodic information both implicitly and explicitly. From the experimental results, our method is better than the previous methods.

In this paper, we address the video anomaly detection (VAD) problem in industrial scenarios. We introduce the first industrial periodic video anomaly detection dataset. This dataset focuses on industrial scenarios and serves as a valuable resource for evaluating VAD methods in industrial environments. Recognizing the inherent periodic characteristics in industrial equipment actions, we propose a novel method that effectively integrates periodic information into a reconstruction-based model. Our approach leverages both implicit and explicit utilization of periodic information to enhance anomaly detection performance. Finally, the incorporation of LoRA adapter and synthetic data pretraining we proposed lead to a substantial reduction in data requirements and model training time, all while ensuring detection accuracy. This advancement significantly accelerates the deployment of VAD tasks in real-world scenarios, facilitating their rapid implementation and efficacy.

% \begin{acks}
% \end{acks}

\bibliographystyle{ACM-Reference-Format}
\bibliography{main}


\begin{thebibliography}{58}

%%% the Web address.

\ifx \showCODEN    \undefined \def \showCODEN     #1{\unskip}     \fi
\ifx \showDOI      \undefined \def \showDOI       #1{#1}\fi
\ifx \showISBNx    \undefined \def \showISBNx     #1{\unskip}     \fi
\ifx \showISBNxiii \undefined \def \showISBNxiii  #1{\unskip}     \fi
\ifx \showISSN     \undefined \def \showISSN      #1{\unskip}     \fi
\ifx \showLCCN     \undefined \def \showLCCN      #1{\unskip}     \fi
\ifx \shownote     \undefined \def \shownote      #1{#1}          \fi
\ifx \showarticletitle \undefined \def \showarticletitle #1{#1}   \fi
\ifx \showURL      \undefined \def \showURL       {\relax}        \fi
% invisible to TeX
\providecommand\bibfield[2]{#2}
\providecommand\bibinfo[2]{#2}
\providecommand\natexlab[1]{#1}
\providecommand\showeprint[2][]{arXiv:#2}

\bibitem[Acsintoae et~al\mbox{.}(2022)]%
        {UBnormal}
\bibfield{author}{\bibinfo{person}{Andra Acsintoae}, \bibinfo{person}{Andrei Florescu}, \bibinfo{person}{Mariana-Iuliana Georgescu}, \bibinfo{person}{Tudor Mare}, \bibinfo{person}{Paul Sumedrea}, \bibinfo{person}{Radu~Tudor Ionescu}, \bibinfo{person}{Fahad~Shahbaz Khan}, {and} \bibinfo{person}{Mubarak Shah}.} \bibinfo{year}{2022}\natexlab{}.
\newblock \showarticletitle{Ubnormal: New benchmark for supervised open-set video anomaly detection}. In \bibinfo{booktitle}{\emph{{CVPR}}}. \bibinfo{pages}{20143--20153}.
\newblock


\bibitem[Adam et~al\mbox{.}(2008)]%
        {Subway}
\bibfield{author}{\bibinfo{person}{Amit Adam}, \bibinfo{person}{Ehud Rivlin}, \bibinfo{person}{Ilan Shimshoni}, {and} \bibinfo{person}{Daviv Reinitz}.} \bibinfo{year}{2008}\natexlab{}.
\newblock \showarticletitle{Robust real-time unusual event detection using multiple fixed-location monitors}.
\newblock \bibinfo{journal}{\emph{IEEE transactions on pattern analysis and machine intelligence}} (\bibinfo{year}{2008}), \bibinfo{pages}{555--560}.
\newblock


\bibitem[Barbalau et~al\mbox{.}(2023)]%
        {barbalau2023ssmtl++}
\bibfield{author}{\bibinfo{person}{Antonio Barbalau}, \bibinfo{person}{Radu~Tudor Ionescu}, \bibinfo{person}{Mariana-Iuliana Georgescu}, \bibinfo{person}{Jacob Dueholm}, \bibinfo{person}{Bharathkumar Ramachandra}, \bibinfo{person}{Kamal Nasrollahi}, \bibinfo{person}{Fahad~Shahbaz Khan}, \bibinfo{person}{Thomas~B Moeslund}, {and} \bibinfo{person}{Mubarak Shah}.} \bibinfo{year}{2023}\natexlab{}.
\newblock \showarticletitle{SSMTL++: Revisiting self-supervised multi-task learning for video anomaly detection}.
\newblock \bibinfo{journal}{\emph{Computer Vision and Image Understanding}} (\bibinfo{year}{2023}), \bibinfo{pages}{103656}.
\newblock


\bibitem[Bergmann et~al\mbox{.}(2019)]%
        {bergmann2019mvtec}
\bibfield{author}{\bibinfo{person}{Paul Bergmann}, \bibinfo{person}{Michael Fauser}, \bibinfo{person}{David Sattlegger}, {and} \bibinfo{person}{Carsten Steger}.} \bibinfo{year}{2019}\natexlab{}.
\newblock \showarticletitle{MVTec AD--A comprehensive real-world dataset for unsupervised anomaly detection}. In \bibinfo{booktitle}{\emph{{CVPR}}}. \bibinfo{pages}{9592--9600}.
\newblock


\bibitem[Bergmann et~al\mbox{.}(2020)]%
        {bergmann2020uninformed}
\bibfield{author}{\bibinfo{person}{Paul Bergmann}, \bibinfo{person}{Michael Fauser}, \bibinfo{person}{David Sattlegger}, {and} \bibinfo{person}{Carsten Steger}.} \bibinfo{year}{2020}\natexlab{}.
\newblock \showarticletitle{Uninformed students: Student-teacher anomaly detection with discriminative latent embeddings}. In \bibinfo{booktitle}{\emph{{CVPR}}}. \bibinfo{pages}{4183--4192}.
\newblock


\bibitem[Cao et~al\mbox{.}(2023)]%
        {NWPU}
\bibfield{author}{\bibinfo{person}{Congqi Cao}, \bibinfo{person}{Yue Lu}, \bibinfo{person}{Peng Wang}, {and} \bibinfo{person}{Yanning Zhang}.} \bibinfo{year}{2023}\natexlab{}.
\newblock \showarticletitle{A New Comprehensive Benchmark for Semi-supervised Video Anomaly Detection and Anticipation}. In \bibinfo{booktitle}{\emph{{CVPR}}}. \bibinfo{pages}{20392--20401}.
\newblock


\bibitem[Carreira and Zisserman(2017)]%
        {I3D}
\bibfield{author}{\bibinfo{person}{Joao Carreira} {and} \bibinfo{person}{Andrew Zisserman}.} \bibinfo{year}{2017}\natexlab{}.
\newblock \showarticletitle{Quo vadis, action recognition? a new model and the kinetics dataset}. In \bibinfo{booktitle}{\emph{{CVPR}}}. \bibinfo{pages}{6299--6308}.
\newblock


\bibitem[Chang et~al\mbox{.}(2020)]%
        {chang2020clustering}
\bibfield{author}{\bibinfo{person}{Yunpeng Chang}, \bibinfo{person}{Zhigang Tu}, \bibinfo{person}{Wei Xie}, {and} \bibinfo{person}{Junsong Yuan}.} \bibinfo{year}{2020}\natexlab{}.
\newblock \showarticletitle{Clustering driven deep autoencoder for video anomaly detection}. In \bibinfo{booktitle}{\emph{{ECCV}}}. \bibinfo{pages}{329--345}.
\newblock


\bibitem[Dosovitskiy et~al\mbox{.}(2017)]%
        {dosovitskiy2017carla}
\bibfield{author}{\bibinfo{person}{Alexey Dosovitskiy}, \bibinfo{person}{German Ros}, \bibinfo{person}{Felipe Codevilla}, \bibinfo{person}{Antonio Lopez}, {and} \bibinfo{person}{Vladlen Koltun}.} \bibinfo{year}{2017}\natexlab{}.
\newblock \showarticletitle{CARLA: An open urban driving simulator}. In \bibinfo{booktitle}{\emph{{CoRL}}}. \bibinfo{pages}{1--16}.
\newblock


\bibitem[Gaidon et~al\mbox{.}(2016)]%
        {gaidon2016virtual}
\bibfield{author}{\bibinfo{person}{Adrien Gaidon}, \bibinfo{person}{Qiao Wang}, \bibinfo{person}{Yohann Cabon}, {and} \bibinfo{person}{Eleonora Vig}.} \bibinfo{year}{2016}\natexlab{}.
\newblock \showarticletitle{Virtual worlds as proxy for multi-object tracking analysis}. In \bibinfo{booktitle}{\emph{{CVPR}}}. \bibinfo{pages}{4340--4349}.
\newblock


\bibitem[Gong et~al\mbox{.}(2019)]%
        {MemAE}
\bibfield{author}{\bibinfo{person}{Dong Gong}, \bibinfo{person}{Lingqiao Liu}, \bibinfo{person}{Vuong Le}, \bibinfo{person}{Budhaditya Saha}, \bibinfo{person}{Moussa~Reda Mansour}, \bibinfo{person}{Svetha Venkatesh}, {and} \bibinfo{person}{Anton van~den Hengel}.} \bibinfo{year}{2019}\natexlab{}.
\newblock \showarticletitle{Memorizing normality to detect anomaly: Memory-augmented deep autoencoder for unsupervised anomaly detection}. In \bibinfo{booktitle}{\emph{{ICCV}}}. \bibinfo{pages}{1705--1714}.
\newblock


\bibitem[Goodfellow et~al\mbox{.}(2014)]%
        {GAN}
\bibfield{author}{\bibinfo{person}{Ian Goodfellow}, \bibinfo{person}{Jean Pouget-Abadie}, \bibinfo{person}{Mehdi Mirza}, \bibinfo{person}{Bing Xu}, \bibinfo{person}{David Warde-Farley}, \bibinfo{person}{Sherjil Ozair}, \bibinfo{person}{Aaron Courville}, {and} \bibinfo{person}{Yoshua Bengio}.} \bibinfo{year}{2014}\natexlab{}.
\newblock \showarticletitle{Generative adversarial nets}.
\newblock \bibinfo{journal}{\emph{Advances in neural information processing systems}}  \bibinfo{volume}{27} (\bibinfo{year}{2014}).
\newblock


\bibitem[Gudovskiy et~al\mbox{.}(2022)]%
        {gudovskiy2022cflow}
\bibfield{author}{\bibinfo{person}{Denis Gudovskiy}, \bibinfo{person}{Shun Ishizaka}, {and} \bibinfo{person}{Kazuki Kozuka}.} \bibinfo{year}{2022}\natexlab{}.
\newblock \showarticletitle{Cflow-ad: Real-time unsupervised anomaly detection with localization via conditional normalizing flows}. In \bibinfo{booktitle}{\emph{{WACV}}}. \bibinfo{pages}{98--107}.
\newblock


\bibitem[Hasan et~al\mbox{.}(2016)]%
        {AE}
\bibfield{author}{\bibinfo{person}{Mahmudul Hasan}, \bibinfo{person}{Jonghyun Choi}, \bibinfo{person}{Jan Neumann}, \bibinfo{person}{Amit~K Roy-Chowdhury}, {and} \bibinfo{person}{Larry~S Davis}.} \bibinfo{year}{2016}\natexlab{}.
\newblock \showarticletitle{Learning temporal regularity in video sequences}. In \bibinfo{booktitle}{\emph{{CVPR}}}. \bibinfo{pages}{733--742}.
\newblock


\bibitem[Hirschorn and Avidan(2023)]%
        {hirschorn2023normalizing}
\bibfield{author}{\bibinfo{person}{Or Hirschorn} {and} \bibinfo{person}{Shai Avidan}.} \bibinfo{year}{2023}\natexlab{}.
\newblock \showarticletitle{Normalizing flows for human pose anomaly detection}. In \bibinfo{booktitle}{\emph{{ICCV}}}. \bibinfo{pages}{13545--13554}.
\newblock


\bibitem[Hu et~al\mbox{.}(2022)]%
        {hu2021lora}
\bibfield{author}{\bibinfo{person}{Edward~J Hu}, \bibinfo{person}{Yelong Shen}, \bibinfo{person}{Phillip Wallis}, \bibinfo{person}{Zeyuan Allen-Zhu}, \bibinfo{person}{Yuanzhi Li}, \bibinfo{person}{Shean Wang}, \bibinfo{person}{Lu Wang}, {and} \bibinfo{person}{Weizhu Chen}.} \bibinfo{year}{2022}\natexlab{}.
\newblock \showarticletitle{Lora: Low-rank adaptation of large language models}.
\newblock  (\bibinfo{year}{2022}).
\newblock


\bibitem[Jackson and Cuzzolin(2021)]%
        {SVD-GAN}
\bibfield{author}{\bibinfo{person}{Samuel~D Jackson} {and} \bibinfo{person}{Fabio Cuzzolin}.} \bibinfo{year}{2021}\natexlab{}.
\newblock \showarticletitle{Svd-gan for real-time unsupervised video anomaly detection}. In \bibinfo{booktitle}{\emph{{BMVC}}}. \bibinfo{pages}{22--25}.
\newblock


\bibitem[Kim et~al\mbox{.}(2021)]%
        {kim2021semi}
\bibfield{author}{\bibinfo{person}{Jin-Hwa Kim}, \bibinfo{person}{Do-Hyeong Kim}, \bibinfo{person}{Saehoon Yi}, {and} \bibinfo{person}{Taehoon Lee}.} \bibinfo{year}{2021}\natexlab{}.
\newblock \showarticletitle{Semi-orthogonal embedding for efficient unsupervised anomaly segmentation}.
\newblock \bibinfo{journal}{\emph{arXiv preprint arXiv:2105.14737}} (\bibinfo{year}{2021}).
\newblock


\bibitem[Le and Kim(2023)]%
        {ASTNet}
\bibfield{author}{\bibinfo{person}{Viet-Tuan Le} {and} \bibinfo{person}{Yong-Guk Kim}.} \bibinfo{year}{2023}\natexlab{}.
\newblock \showarticletitle{Attention-based residual autoencoder for video anomaly detection}.
\newblock \bibinfo{journal}{\emph{Applied Intelligence}} \bibinfo{volume}{53}, \bibinfo{number}{3} (\bibinfo{year}{2023}), \bibinfo{pages}{3240--3254}.
\newblock


\bibitem[Li et~al\mbox{.}(2021b)]%
        {li2021cutpaste}
\bibfield{author}{\bibinfo{person}{Chun-Liang Li}, \bibinfo{person}{Kihyuk Sohn}, \bibinfo{person}{Jinsung Yoon}, {and} \bibinfo{person}{Tomas Pfister}.} \bibinfo{year}{2021}\natexlab{b}.
\newblock \showarticletitle{Cutpaste: Self-supervised learning for anomaly detection and localization}. In \bibinfo{booktitle}{\emph{{CVPR}}}. \bibinfo{pages}{9664--9674}.
\newblock


\bibitem[Li et~al\mbox{.}(2021a)]%
        {li2021variational}
\bibfield{author}{\bibinfo{person}{Jing Li}, \bibinfo{person}{Qingwang Huang}, \bibinfo{person}{Yingjun Du}, \bibinfo{person}{Xiantong Zhen}, \bibinfo{person}{Shengyong Chen}, {and} \bibinfo{person}{Ling Shao}.} \bibinfo{year}{2021}\natexlab{a}.
\newblock \showarticletitle{Variational abnormal behavior detection with motion consistency}.
\newblock \bibinfo{journal}{\emph{IEEE Transactions on Image Processing}} (\bibinfo{year}{2021}), \bibinfo{pages}{275--286}.
\newblock


\bibitem[Li et~al\mbox{.}(2013)]%
        {UCSDped}
\bibfield{author}{\bibinfo{person}{Weixin Li}, \bibinfo{person}{Vijay Mahadevan}, {and} \bibinfo{person}{Nuno Vasconcelos}.} \bibinfo{year}{2013}\natexlab{}.
\newblock \showarticletitle{Anomaly detection and localization in crowded scenes}.
\newblock \bibinfo{journal}{\emph{IEEE transactions on pattern analysis and machine intelligence}} (\bibinfo{year}{2013}), \bibinfo{pages}{18--32}.
\newblock


\bibitem[Liu et~al\mbox{.}(2023a)]%
        {DMAD}
\bibfield{author}{\bibinfo{person}{Wenrui Liu}, \bibinfo{person}{Hong Chang}, \bibinfo{person}{Bingpeng Ma}, \bibinfo{person}{Shiguang Shan}, {and} \bibinfo{person}{Xilin Chen}.} \bibinfo{year}{2023}\natexlab{a}.
\newblock \showarticletitle{Diversity-measurable anomaly detection}. In \bibinfo{booktitle}{\emph{{CVPR}}}. \bibinfo{pages}{12147--12156}.
\newblock


\bibitem[Liu et~al\mbox{.}(2020)]%
        {liu2020towards}
\bibfield{author}{\bibinfo{person}{Wenqian Liu}, \bibinfo{person}{Runze Li}, \bibinfo{person}{Meng Zheng}, \bibinfo{person}{Srikrishna Karanam}, \bibinfo{person}{Ziyan Wu}, \bibinfo{person}{Bir Bhanu}, \bibinfo{person}{Richard~J Radke}, {and} \bibinfo{person}{Octavia Camps}.} \bibinfo{year}{2020}\natexlab{}.
\newblock \showarticletitle{Towards visually explaining variational autoencoders}. In \bibinfo{booktitle}{\emph{{CVPR}}}. \bibinfo{pages}{8642--8651}.
\newblock


\bibitem[Liu et~al\mbox{.}(2018)]%
        {liu2018future}
\bibfield{author}{\bibinfo{person}{Wen Liu}, \bibinfo{person}{Weixin Luo}, \bibinfo{person}{Dongze Lian}, {and} \bibinfo{person}{Shenghua Gao}.} \bibinfo{year}{2018}\natexlab{}.
\newblock \showarticletitle{Future frame prediction for anomaly detection--a new baseline}. In \bibinfo{booktitle}{\emph{{CVPR}}}. \bibinfo{pages}{6536--6545}.
\newblock


\bibitem[Liu et~al\mbox{.}(2023b)]%
        {causality-inspired}
\bibfield{author}{\bibinfo{person}{Yang Liu}, \bibinfo{person}{Zhaoyang Xia}, \bibinfo{person}{Mengyang Zhao}, \bibinfo{person}{Donglai Wei}, \bibinfo{person}{Yuzheng Wang}, \bibinfo{person}{Siao Liu}, \bibinfo{person}{Bobo Ju}, \bibinfo{person}{Gaoyun Fang}, \bibinfo{person}{Jing Liu}, {and} \bibinfo{person}{Liang Song}.} \bibinfo{year}{2023}\natexlab{b}.
\newblock \showarticletitle{Learning causality-inspired representation consistency for video anomaly detection}. In \bibinfo{booktitle}{\emph{{ACM MM}}}. \bibinfo{pages}{203--212}.
\newblock


\bibitem[Liu et~al\mbox{.}(2021b)]%
        {liu2021unsupervised}
\bibfield{author}{\bibinfo{person}{Yunfei Liu}, \bibinfo{person}{Chaoqun Zhuang}, {and} \bibinfo{person}{Feng Lu}.} \bibinfo{year}{2021}\natexlab{b}.
\newblock \showarticletitle{Unsupervised two-stage anomaly detection}.
\newblock \bibinfo{journal}{\emph{arXiv preprint arXiv:2103.11671}} (\bibinfo{year}{2021}).
\newblock


\bibitem[Liu et~al\mbox{.}(2021a)]%
        {liu2021hybrid}
\bibfield{author}{\bibinfo{person}{Zhian Liu}, \bibinfo{person}{Yongwei Nie}, \bibinfo{person}{Chengjiang Long}, \bibinfo{person}{Qing Zhang}, {and} \bibinfo{person}{Guiqing Li}.} \bibinfo{year}{2021}\natexlab{a}.
\newblock \showarticletitle{A hybrid video anomaly detection framework via memory-augmented flow reconstruction and flow-guided frame prediction}. In \bibinfo{booktitle}{\emph{{ICCV}}}. \bibinfo{pages}{13588--13597}.
\newblock


\bibitem[Liu et~al\mbox{.}(2022)]%
        {video-swin-T}
\bibfield{author}{\bibinfo{person}{Ze Liu}, \bibinfo{person}{Jia Ning}, \bibinfo{person}{Yue Cao}, \bibinfo{person}{Yixuan Wei}, \bibinfo{person}{Zheng Zhang}, \bibinfo{person}{Stephen Lin}, {and} \bibinfo{person}{Han Hu}.} \bibinfo{year}{2022}\natexlab{}.
\newblock \showarticletitle{Video swin transformer}. In \bibinfo{booktitle}{\emph{{CVPR}}}. \bibinfo{pages}{3202--3211}.
\newblock


\bibitem[Lu et~al\mbox{.}(2013)]%
        {CUHK}
\bibfield{author}{\bibinfo{person}{Cewu Lu}, \bibinfo{person}{Jianping Shi}, {and} \bibinfo{person}{Jiaya Jia}.} \bibinfo{year}{2013}\natexlab{}.
\newblock \showarticletitle{Abnormal event detection at 150 fps in matlab}. In \bibinfo{booktitle}{\emph{{ICCV}}}. \bibinfo{pages}{2720--2727}.
\newblock


\bibitem[Lu et~al\mbox{.}(2020)]%
        {lu2020few}
\bibfield{author}{\bibinfo{person}{Yiwei Lu}, \bibinfo{person}{Frank Yu}, \bibinfo{person}{Mahesh Kumar~Krishna Reddy}, {and} \bibinfo{person}{Yang Wang}.} \bibinfo{year}{2020}\natexlab{}.
\newblock \showarticletitle{Few-shot scene-adaptive anomaly detection}. In \bibinfo{booktitle}{\emph{{ECCV}}}. \bibinfo{pages}{125--141}.
\newblock


\bibitem[Luo et~al\mbox{.}(2017)]%
        {ShangHaiTech}
\bibfield{author}{\bibinfo{person}{Weixin Luo}, \bibinfo{person}{Wen Liu}, {and} \bibinfo{person}{Shenghua Gao}.} \bibinfo{year}{2017}\natexlab{}.
\newblock \showarticletitle{A revisit of sparse coding based anomaly detection in stacked rnn framework}. In \bibinfo{booktitle}{\emph{{ICCV}}}. \bibinfo{pages}{341--349}.
\newblock


\bibitem[Mishra et~al\mbox{.}(2021)]%
        {mishra2021btad}
\bibfield{author}{\bibinfo{person}{Pankaj Mishra}, \bibinfo{person}{Riccardo Verk}, \bibinfo{person}{Daniele Fornasier}, \bibinfo{person}{Claudio Piciarelli}, {and} \bibinfo{person}{Gian~Luca Foresti}.} \bibinfo{year}{2021}\natexlab{}.
\newblock \showarticletitle{VT-ADL: A vision transformer network for image anomaly detection and localization}. In \bibinfo{booktitle}{\emph{ISIE}}. \bibinfo{pages}{01--06}.
\newblock


\bibitem[of~Minnesota({[n.\,d.]})]%
        {UMN}
\bibfield{author}{\bibinfo{person}{University of Minnesota}.} \bibinfo{year}{[n.\,d.]}\natexlab{}.
\newblock \showarticletitle{Unusual crowd activity dataset of university of minnesota}.
\newblock \bibinfo{journal}{\emph{\url{http://mha.cs.umn.edu/proj_events.shtml##crowd}}} (\bibinfo{year}{[n.\,d.]}).
\newblock


\bibitem[Pirnay and Chai(2022)]%
        {pirnay2022inpainting}
\bibfield{author}{\bibinfo{person}{Jonathan Pirnay} {and} \bibinfo{person}{Keng Chai}.} \bibinfo{year}{2022}\natexlab{}.
\newblock \showarticletitle{Inpainting transformer for anomaly detection}. In \bibinfo{booktitle}{\emph{{ICIAP}}}. \bibinfo{pages}{394--406}.
\newblock


\bibitem[Ramachandra and Jones(2020)]%
        {Street}
\bibfield{author}{\bibinfo{person}{Bharathkumar Ramachandra} {and} \bibinfo{person}{Michael Jones}.} \bibinfo{year}{2020}\natexlab{}.
\newblock \showarticletitle{Street scene: A new dataset and evaluation protocol for video anomaly detection}. In \bibinfo{booktitle}{\emph{{WACV}}}. \bibinfo{pages}{2569--2578}.
\newblock


\bibitem[Reiss and Hoshen(2022)]%
        {reiss2022attribute}
\bibfield{author}{\bibinfo{person}{Tal Reiss} {and} \bibinfo{person}{Yedid Hoshen}.} \bibinfo{year}{2022}\natexlab{}.
\newblock \showarticletitle{Attribute-based Representations for Accurate and Interpretable Video Anomaly Detection}.
\newblock \bibinfo{journal}{\emph{arXiv preprint arXiv:2212.00789}} (\bibinfo{year}{2022}).
\newblock


\bibitem[Rippel et~al\mbox{.}(2022)]%
        {rippel2021transfer}
\bibfield{author}{\bibinfo{person}{Oliver Rippel}, \bibinfo{person}{Arnav Chavan}, \bibinfo{person}{Chucai Lei}, {and} \bibinfo{person}{Dorit Merhof}.} \bibinfo{year}{2022}\natexlab{}.
\newblock \showarticletitle{Transfer learning gaussian anomaly detection by fine-tuning representations}.
\newblock  (\bibinfo{year}{2022}), \bibinfo{pages}{45--56}.
\newblock


\bibitem[Rodrigues et~al\mbox{.}(2020)]%
        {IITB}
\bibfield{author}{\bibinfo{person}{Royston Rodrigues}, \bibinfo{person}{Neha Bhargava}, \bibinfo{person}{Rajbabu Velmurugan}, {and} \bibinfo{person}{Subhasis Chaudhuri}.} \bibinfo{year}{2020}\natexlab{}.
\newblock \showarticletitle{Multi-timescale trajectory prediction for abnormal human activity detection}. In \bibinfo{booktitle}{\emph{{WACV}}}. \bibinfo{pages}{2626--2634}.
\newblock


\bibitem[Salehi et~al\mbox{.}(2021)]%
        {salehi2021multiresolution}
\bibfield{author}{\bibinfo{person}{Mohammadreza Salehi}, \bibinfo{person}{Niousha Sadjadi}, \bibinfo{person}{Soroosh Baselizadeh}, \bibinfo{person}{Mohammad~H Rohban}, {and} \bibinfo{person}{Hamid~R Rabiee}.} \bibinfo{year}{2021}\natexlab{}.
\newblock \showarticletitle{Multiresolution knowledge distillation for anomaly detection}. In \bibinfo{booktitle}{\emph{{CVPR}}}. \bibinfo{pages}{14902--14912}.
\newblock


\bibitem[Schlegl et~al\mbox{.}(2017)]%
        {schlegl2017unsupervised}
\bibfield{author}{\bibinfo{person}{Thomas Schlegl}, \bibinfo{person}{Philipp Seeb{\"o}ck}, \bibinfo{person}{Sebastian~M Waldstein}, \bibinfo{person}{Ursula Schmidt-Erfurth}, {and} \bibinfo{person}{Georg Langs}.} \bibinfo{year}{2017}\natexlab{}.
\newblock \showarticletitle{Unsupervised anomaly detection with generative adversarial networks to guide marker discovery}. In \bibinfo{booktitle}{\emph{{IPMI}}}. \bibinfo{pages}{146--157}.
\newblock


\bibitem[Sultani et~al\mbox{.}(2018)]%
        {sultani2018real}
\bibfield{author}{\bibinfo{person}{Waqas Sultani}, \bibinfo{person}{Chen Chen}, {and} \bibinfo{person}{Mubarak Shah}.} \bibinfo{year}{2018}\natexlab{}.
\newblock \showarticletitle{Real-world anomaly detection in surveillance videos}. In \bibinfo{booktitle}{\emph{{CVPR}}}. \bibinfo{pages}{6479--6488}.
\newblock


\bibitem[Sun and Gong(2023)]%
        {sun2023hierarchical}
\bibfield{author}{\bibinfo{person}{Shengyang Sun} {and} \bibinfo{person}{Xiaojin Gong}.} \bibinfo{year}{2023}\natexlab{}.
\newblock \showarticletitle{Hierarchical Semantic Contrast for Scene-aware Video Anomaly Detection}. In \bibinfo{booktitle}{\emph{{CVPR}}}. \bibinfo{pages}{22846--22856}.
\newblock


\bibitem[Tan et~al\mbox{.}(2021)]%
        {tan2021trustmae}
\bibfield{author}{\bibinfo{person}{Daniel~Stanley Tan}, \bibinfo{person}{Yi-Chun Chen}, \bibinfo{person}{Trista Pei-Chun Chen}, {and} \bibinfo{person}{Wei-Chao Chen}.} \bibinfo{year}{2021}\natexlab{}.
\newblock \showarticletitle{Trustmae: A noise-resilient defect classification framework using memory-augmented auto-encoders with trust regions}. In \bibinfo{booktitle}{\emph{{WACV}}}. \bibinfo{pages}{276--285}.
\newblock


\bibitem[Tao et~al\mbox{.}(2022)]%
        {tao2022deep}
\bibfield{author}{\bibinfo{person}{Xian Tao}, \bibinfo{person}{Xinyi Gong}, \bibinfo{person}{Xin Zhang}, \bibinfo{person}{Shaohua Yan}, {and} \bibinfo{person}{Chandranath Adak}.} \bibinfo{year}{2022}\natexlab{}.
\newblock \showarticletitle{Deep learning for unsupervised anomaly localization in industrial images: A survey}.
\newblock \bibinfo{journal}{\emph{IEEE Transactions on Instrumentation and Measurement}} (\bibinfo{year}{2022}).
\newblock


\bibitem[Tsai et~al\mbox{.}(2022)]%
        {tsai2022multi}
\bibfield{author}{\bibinfo{person}{Chin-Chia Tsai}, \bibinfo{person}{Tsung-Hsuan Wu}, {and} \bibinfo{person}{Shang-Hong Lai}.} \bibinfo{year}{2022}\natexlab{}.
\newblock \showarticletitle{Multi-scale patch-based representation learning for image anomaly detection and segmentation}. In \bibinfo{booktitle}{\emph{{WACV}}}. \bibinfo{pages}{3992--4000}.
\newblock


\bibitem[Wang et~al\mbox{.}(2024)]%
        {Real-IAD}
\bibfield{author}{\bibinfo{person}{Chengjie Wang}, \bibinfo{person}{Wenbing Zhu}, \bibinfo{person}{Bin-Bin Gao}, \bibinfo{person}{Zhenye Gan}, \bibinfo{person}{Jianning Zhang}, \bibinfo{person}{Zhihao Gu}, \bibinfo{person}{Shuguang Qian}, \bibinfo{person}{Mingang Chen}, {and} \bibinfo{person}{Lizhuang Ma}.} \bibinfo{year}{2024}\natexlab{}.
\newblock \showarticletitle{Real-IAD: A Real-World Multi-View Dataset for Benchmarking Versatile Industrial Anomaly Detection}.
\newblock \bibinfo{journal}{\emph{arXiv preprint arXiv:2403.12580}} (\bibinfo{year}{2024}).
\newblock


\bibitem[Wang et~al\mbox{.}(2022)]%
        {Jigsaw}
\bibfield{author}{\bibinfo{person}{Guodong Wang}, \bibinfo{person}{Yunhong Wang}, \bibinfo{person}{Jie Qin}, \bibinfo{person}{Dongming Zhang}, \bibinfo{person}{Xiuguo Bao}, {and} \bibinfo{person}{Di Huang}.} \bibinfo{year}{2022}\natexlab{}.
\newblock \showarticletitle{Video anomaly detection by solving decoupled spatio-temporal jigsaw puzzles}. In \bibinfo{booktitle}{\emph{{ECCV}}}. \bibinfo{pages}{494--511}.
\newblock


\bibitem[Wang et~al\mbox{.}(2020)]%
        {wang2020image}
\bibfield{author}{\bibinfo{person}{Lu Wang}, \bibinfo{person}{Dongkai Zhang}, \bibinfo{person}{Jiahao Guo}, {and} \bibinfo{person}{Yuexing Han}.} \bibinfo{year}{2020}\natexlab{}.
\newblock \showarticletitle{Image anomaly detection using normal data only by latent space resampling}.
\newblock \bibinfo{journal}{\emph{Applied Sciences}} \bibinfo{number}{23} (\bibinfo{year}{2020}), \bibinfo{pages}{8660}.
\newblock


\bibitem[Xiao et~al\mbox{.}(2023)]%
        {xiao2023divide}
\bibfield{author}{\bibinfo{person}{Jian Xiao}, \bibinfo{person}{Tianyuan Liu}, {and} \bibinfo{person}{Genlin Ji}.} \bibinfo{year}{2023}\natexlab{}.
\newblock \showarticletitle{Divide and Conquer in Video Anomaly Detection: A Comprehensive Review and New Approach}.
\newblock \bibinfo{journal}{\emph{arXiv preprint arXiv:2309.14622}} (\bibinfo{year}{2023}).
\newblock


\bibitem[Yan et~al\mbox{.}(2021)]%
        {yan2021unsupervised}
\bibfield{author}{\bibinfo{person}{Yi Yan}, \bibinfo{person}{Deming Wang}, \bibinfo{person}{Guangliang Zhou}, {and} \bibinfo{person}{Qijun Chen}.} \bibinfo{year}{2021}\natexlab{}.
\newblock \showarticletitle{Unsupervised anomaly segmentation via multilevel image reconstruction and adaptive attention-level transition}.
\newblock \bibinfo{journal}{\emph{IEEE Transactions on Instrumentation and Measurement}} (\bibinfo{year}{2021}), \bibinfo{pages}{1--12}.
\newblock


\bibitem[Yu et~al\mbox{.}(2021)]%
        {yu2021fastflow}
\bibfield{author}{\bibinfo{person}{Jiawei Yu}, \bibinfo{person}{Ye Zheng}, \bibinfo{person}{Xiang Wang}, \bibinfo{person}{Wei Li}, \bibinfo{person}{Yushuang Wu}, \bibinfo{person}{Rui Zhao}, {and} \bibinfo{person}{Liwei Wu}.} \bibinfo{year}{2021}\natexlab{}.
\newblock \showarticletitle{Fastflow: Unsupervised anomaly detection and localization via 2d normalizing flows}.
\newblock \bibinfo{journal}{\emph{arXiv preprint arXiv:2111.07677}} (\bibinfo{year}{2021}).
\newblock


\bibitem[Zaheer et~al\mbox{.}(2020)]%
        {zaheer2020claws}
\bibfield{author}{\bibinfo{person}{Muhammad~Zaigham Zaheer}, \bibinfo{person}{Arif Mahmood}, \bibinfo{person}{Marcella Astrid}, {and} \bibinfo{person}{Seung-Ik Lee}.} \bibinfo{year}{2020}\natexlab{}.
\newblock \showarticletitle{Claws: Clustering assisted weakly supervised learning with normalcy suppression for anomalous event detection}. In \bibinfo{booktitle}{\emph{{ECCV}}}. \bibinfo{pages}{358--376}.
\newblock


\bibitem[Zaigham~Zaheer et~al\mbox{.}(2021)]%
        {zaigham2021cleaning}
\bibfield{author}{\bibinfo{person}{Muhammad Zaigham~Zaheer}, \bibinfo{person}{Jin-ha Lee}, \bibinfo{person}{Marcella Astrid}, \bibinfo{person}{Arif Mahmood}, {and} \bibinfo{person}{Seung-Ik Lee}.} \bibinfo{year}{2021}\natexlab{}.
\newblock \showarticletitle{Cleaning Label Noise with Clusters for Minimally Supervised Anomaly Detection}.
\newblock \bibinfo{journal}{\emph{arXiv preprint arXiv:2104.14770}} (\bibinfo{year}{2021}).
\newblock


\bibitem[Zhang et~al\mbox{.}(2023)]%
        {Anomaly-Segmentation}
\bibfield{author}{\bibinfo{person}{Ji Zhang}, \bibinfo{person}{Xiao Wu}, \bibinfo{person}{Zhi-Qi Cheng}, \bibinfo{person}{Qi He}, {and} \bibinfo{person}{Wei Li}.} \bibinfo{year}{2023}\natexlab{}.
\newblock \showarticletitle{Improving Anomaly Segmentation with Multi-Granularity Cross-Domain Alignment}. In \bibinfo{booktitle}{\emph{{ACM MM}}}. \bibinfo{pages}{8515--8524}.
\newblock


\bibitem[Zhong et~al\mbox{.}(2019)]%
        {zhong2019graph}
\bibfield{author}{\bibinfo{person}{Jia-Xing Zhong}, \bibinfo{person}{Nannan Li}, \bibinfo{person}{Weijie Kong}, \bibinfo{person}{Shan Liu}, \bibinfo{person}{Thomas~H Li}, {and} \bibinfo{person}{Ge Li}.} \bibinfo{year}{2019}\natexlab{}.
\newblock \showarticletitle{Graph convolutional label noise cleaner: Train a plug-and-play action classifier for anomaly detection}. In \bibinfo{booktitle}{\emph{{CVPR}}}. \bibinfo{pages}{1237--1246}.
\newblock


\bibitem[Zhu et~al\mbox{.}(2023)]%
        {zhu2023cross}
\bibfield{author}{\bibinfo{person}{Dongliang Zhu}, \bibinfo{person}{Ruimin Hu}, \bibinfo{person}{Shengli Song}, \bibinfo{person}{Xiang Guo}, \bibinfo{person}{Xixi Li}, {and} \bibinfo{person}{Zheng Wang}.} \bibinfo{year}{2023}\natexlab{}.
\newblock \showarticletitle{Cross-Illumination Video Anomaly Detection Benchmark}. In \bibinfo{booktitle}{\emph{{ACM MM}}}. \bibinfo{pages}{2516--2525}.
\newblock


\bibitem[Zou et~al\mbox{.}(2021)]%
        {zou2021progressive}
\bibfield{author}{\bibinfo{person}{Xueyan Zou}, \bibinfo{person}{Linjie Yang}, \bibinfo{person}{Ding Liu}, {and} \bibinfo{person}{Yong~Jae Lee}.} \bibinfo{year}{2021}\natexlab{}.
\newblock \showarticletitle{Progressive temporal feature alignment network for video inpainting}. In \bibinfo{booktitle}{\emph{{CVPR}}}. \bibinfo{pages}{16448--16457}.
\newblock


\end{thebibliography}

% \appendix









\end{document}